\pdfoutput=1
\documentclass[11pt]{article}

\usepackage[]{ACL2023}

\usepackage{times}
\usepackage{latexsym}

\usepackage{amsmath,amsfonts,bm}

\def\eqref#1{equation~\ref{#1}}
\def\1{\bm{1}}

\def\vv{{\bm{v}}}

\DeclareMathAlphabet{\mathsfit}{\encodingdefault}{\sfdefault}{m}{sl}
\SetMathAlphabet{\mathsfit}{bold}{\encodingdefault}{\sfdefault}{bx}{n}
\newcommand{\tens}[1]{\bm{\mathsfit{#1}}}

\def\sS{{\mathbb{S}}}

\newcommand{\etens}[1]{\mathsfit{#1}}

\usepackage{times}
\usepackage{latexsym}
\usepackage{subcaption}
\usepackage{caption}
\usepackage{graphicx}
\usepackage{graphics}
\usepackage{multirow}
\usepackage{booktabs}
\usepackage{colortbl}
\usepackage{xcolor}
\usepackage{amsmath}
\usepackage{amssymb}
\usepackage{amsfonts}
\usepackage{algorithm}
\usepackage{algorithmic}
\usepackage{float}
\usepackage{bbm}
\usepackage{mathtools}
\usepackage{booktabs}
\usepackage[utf8]{inputenc}
\usepackage{enumitem}
\usepackage{amsthm}
\usepackage{inconsolata}
\usepackage{microtype}
\usepackage[T1]{fontenc}

\usepackage[T1]{fontenc}
\usepackage[utf8]{inputenc}

\usepackage{microtype}

\usepackage{inconsolata}
\let\SUP\textsuperscript
\newcommand\yeon[1]{\textcolor{black}{{#1}}}
\newcommand\revisionyeon[1]{\textcolor{black}{{#1}}}

\title{Ranking-Enhanced Unsupervised Sentence Representation Learning}

\author{{Yeon Seonwoo\SUP{$\dagger$ *}{\normalfont ,}}
{Guoyin Wang\SUP{$\ddagger$}{\normalfont,}}
{Changmin Seo\SUP{$\ddagger$}{\normalfont,}}
{Sajal Choudhary\SUP{$\ddagger$}{\normalfont,}}\\
{{\bf Jiwei Li} \SUP{$\mathsection$}{\normalfont,}}
{{\bf Xiang Li} \SUP{$\ddagger$}{\normalfont,}}
{{\bf Puyang Xu} \SUP{$\ddagger$}{\normalfont,}}
{{\bf Sunghyun Park} \SUP{$\ddagger$}{\normalfont,}}
{{ \bf Alice Oh} \SUP{$\dagger$}{\normalfont}}
\\
\SUP{$\dagger$}KAIST,
\SUP{$\ddagger$}Amazon, 
\SUP{$\mathsection$}Zhejiang Univeristy\\
\tt{yeon.seonwoo@kaist.ac.kr}\\
\tt{\{guoyiwan, changmis, sajalc, lixxiang, puyax, sunghyu\}@amazon.com}\\
\tt{jiwei\_li@zju.edu.cn}\\
\tt{alice.oh@kaist.edu}
}

\begin{document}
\maketitle
\def\thefootnote{*}\footnotetext{This work was done during an internship at Amazon.}
\renewcommand{\thefootnote}{\arabic{footnote}}
\begin{abstract}
\yeon{Unsupervised sentence representation learning has progressed through contrastive learning and data augmentation methods such as dropout masking.
Despite this progress, sentence encoders are still limited to using only an input sentence when predicting its semantic vector.
In this work, we show that the semantic meaning of a sentence is also determined by nearest-neighbor sentences that are similar to the input sentence.
Based on this finding, we propose a novel unsupervised sentence encoder, \textbf{RankEncoder}.
RankEncoder predicts the semantic vector of an input sentence by leveraging its relationship with other sentences in an external corpus, as well as the input sentence itself.
We evaluate RankEncoder on semantic textual benchmark datasets.
From the experimental results, we verify that 1) RankEncoder achieves 80.07\% Spearman's correlation, a 1.1\% absolute improvement compared to the previous state-of-the-art performance, 2) RankEncoder is universally applicable to existing unsupervised sentence embedding methods, and 3) RankEncoder is specifically effective for predicting the similarity scores of similar sentence pairs.\footnote{We provide the implementation of RankEncoder at \url{https://github.com/yeonsw/RankEncoder.git}}
}
\end{abstract}

\section{Introduction}
\yeon{Sentence representation learning aims to encode sentences into a semantic vector space.
This task has been a fundamental task in natural language processing (NLP), as universal sentence vectors are widely applicable to many NLP tasks \cite{kiros2015skip, hill2016learning, conneau2017supervised, logeswaran2018efficient, cer2018universal, reimers2019sentence}.
Recently, unsupervised sentence embedding methods have arisen as they have shown a potential to overcome limited labeled data with simple data augmentation methods~\citep{gao-etal-2021-simcse, wang2022sncse, yan2021consert, liu2021fast, wu2021esimcse, izacard2021unsupervised, kim2021self}.
These approaches minimize the distance between the vector representations of similar sentences, called positive pairs, while maximizing the distance between those of dissimilar sentences, called negative pairs.
Many studies have focused on developing better positive and negative pair sampling methods.
Data augmentation methods such as dropout masking~\citep{gao-etal-2021-simcse}, token shuffling~\citep{yan2021consert}, and sentence negation~\citep{wang2022sncse} have been proposed and achieved comparable semantic textual similarity performance to sentence encoders trained on human-annotated datasets.
}

\begin{figure}[t]
    \begin{center}
    \includegraphics[width=0.99\linewidth]{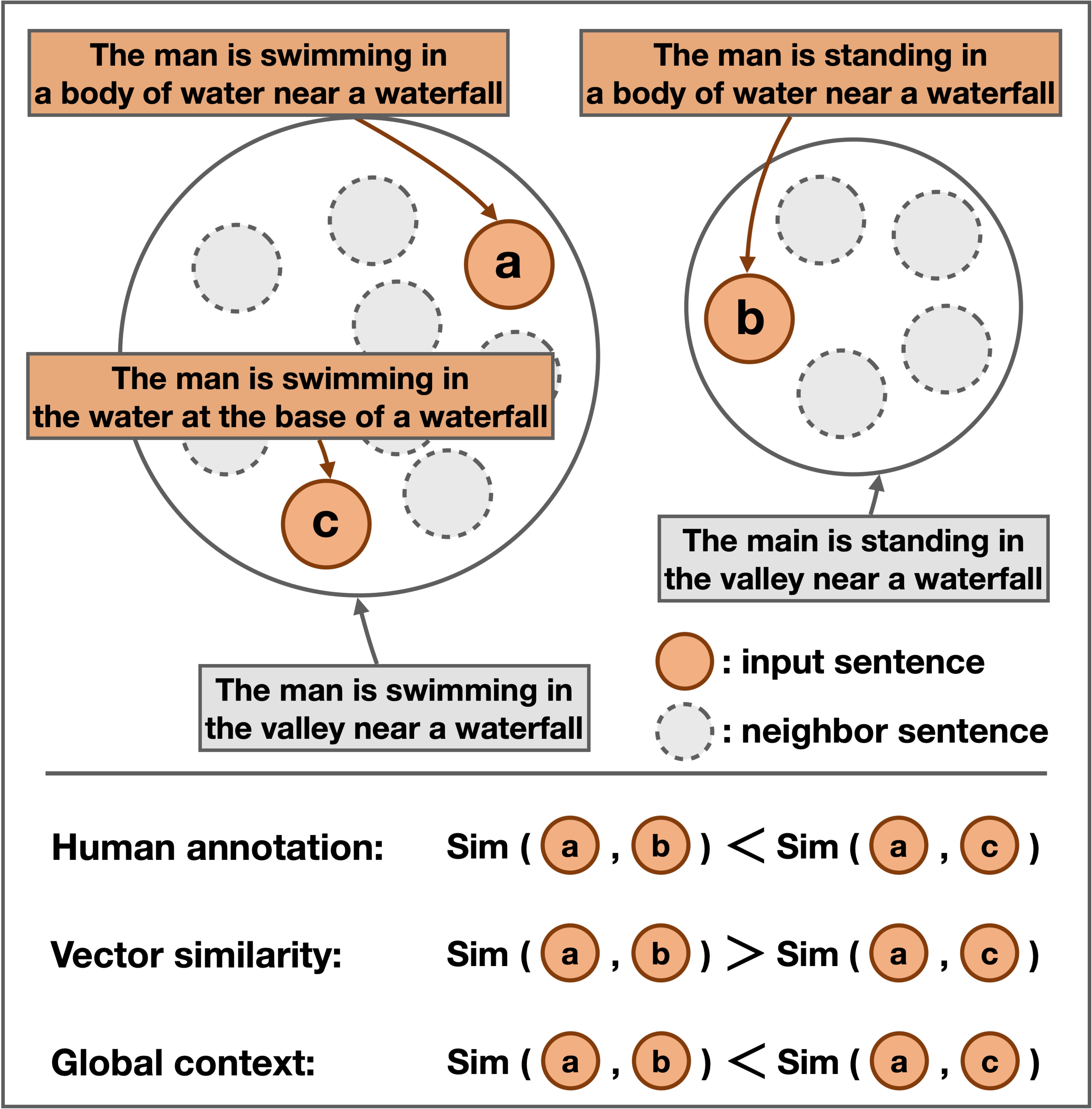}
    \end{center}
    \vspace{-0.5em}
    \caption{
    Vector representations of sentences and their neighbor sentences. The neighbor sentences reveal that $(a, c)$ share more semantic meanings than $(a, b)$. This captures more accurate semantic similarity scores than their vectors.
    }
    \label{fig:intro_example}
    \vspace{-1em}
\end{figure}

\yeon{The semantic meaning of a sentence is not only determined by the words within the sentence itself but also by other sentences with similar meanings.
However, previous unsupervised sentence embedding methods use only the input sentence when predicting its semantic vector.
Figure \ref{fig:intro_example} shows example sentences and their semantic vector space.
In this figure, the human-annotated similarity scores indicate that sentence pair $(a, c)$ is more similar than $(a, b)$.
However, the similarity scores computed by their sentence vectors indicate the opposite result; the vector representations of $a$ and $b$ are closer than $a$ and $c$ as they have more overlapping words than $a$ and $c$.
This problem can be alleviated by leveraging the distance between the input sentence and other sentences in a large corpus.
The vectors of their neighbor sentences approximate the overall semantic distribution, and the semantic distribution reveals that sentences $a$ and $c$ are likely to share more semantic meanings than sentences $a$ and $b$.
This facilitates the accurate prediction of semantic similarity between sentences.}

\yeon{In this paper, we propose \textbf{RankEncoder}, a novel unsupervised sentence encoder that leverages a large number of sentences in an external corpus.
For a given corpus with $n$ number of sentences and an input sentence, RankEncoder computes a rank vector, an $n$-dimensional vector in which $i$'th element represents the distance between the input sentence and $i$'th sentence in the corpus; RankEncoder uses an existing unsupervised sentence encoder, $E$, to compute the distances.
Then, two sentences that share the same neighbor sentences (e.g., sentence $a$ and $c$ in Fig \ref{fig:intro_example}) have similar rank vector representations.
We verify that using the similarity between rank vectors captures better semantic similarity than their vector representation computed by the base encoder, $E$, (Fig \ref{fig:inc}) without further training.
We further leverage the similarity scores predicted by the rank vectors to train another sentence encoder and achieve a better sentence encoder (Table \ref{tab:sts}).
}

\yeon{From experiments on seven STS benchmark datasets, we verify that 1) rank vectors are effective for capturing the semantic similarity of similar sentences, 2) RankEncoder is applicable to any unsupervised sentence encoders, resulting in performance improvement, and 3) this improvement is also valid for the previous state-of-the-art sentence encoder and leads to a new state-of-the-art semantic textual similarity performance.
First, we measure the performance of RankEncoder and the baselines on three sentence pair groups divided by their similarity scores.
The experimental results show that RankEncoder is effective on similar sentence pairs.
Second, we apply RankEncoder to the three unsupervised sentence encoders, SimCSE~\citep{gao-etal-2021-simcse}, PromptBERT~\citep{jiang2022promptbert}, and SNCSE~\citep{wang2022sncse}, then verify that our approach brings performance improvement to each encoder.
Third, we apply RankEncoder to the state-of-the-art unsupervised sentence encoder~\citep{wang2022sncse} and achieve a 1.1\% improvement; the previous state-of-the-art is 78.97 Spearman's correlation, and we achieve 80.07 Spearman's correlation.
}

\yeon{The contributions of this paper are three folds. 
First, we demonstrate that the semantic meaning of a sentence is also determined by its nearest-neighbor sentences as well as the words within the sentence itself.
Second, we propose RankEncoder, which leverages a large number of sentences to capture the semantic meanings of sentences.
Third, we achieve state-of-the-art STS performance and reduce the gap between supervised and unsupervised sentence encoders; the performances of our method and the state-of-the-art supervised sentence encoder~\citep{jiang2022promptbert} are 80.07 and 81.97, respectively.}
\section{Related Works}
\begin{figure*}[t]
    \centering
    \includegraphics[width=1.0\linewidth]{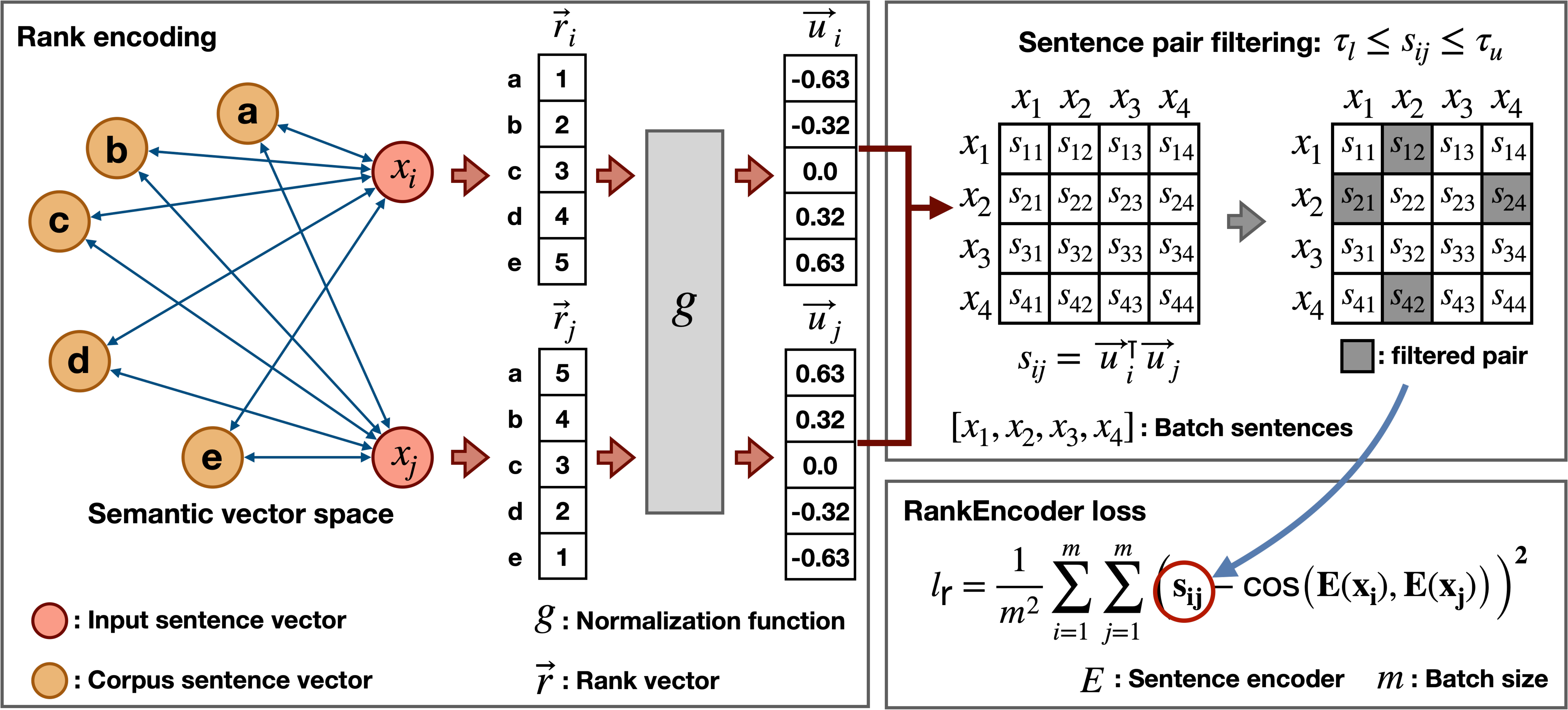}
    \caption{The overall illustration of RankEncoder. The left figure shows the process for computing rank vectors. For a given sentence pair, $(x_i, x_j)$, RankEncoder computes orders of sentences in the corpus by their similarity scores to the input sentence and normalizes these orders with the function $g$; the similarity scores are computed by the base encoder, $E_1$. The right figures show the training process of RankEncoder. For a batch of sentences, RankEncoder computes similarity scores of sentences with their rank vectors and trains the sentence encoder with the mean square error of these scores.}
    \label{fig:method}
    \vspace{-1em}
\end{figure*}

\yeon{Unsupervised sentence representation learning has progressed through contrastive learning with positive and negative sentence pair sampling methods \cite{gao-etal-2021-simcse, jiang2022promptbert, chuang2022diffcse, wang2022sncse}.
SimCSE~\citep{gao-etal-2021-simcse} and ConSERT~\citep{yan2021consert} apply data augmentation methods such as dropout masking, token shuffling, and adversarial attacks to an input sentence and sample a positive pair.
However, these data augmentation methods often change the meaning of the input sentence and generate dissimilar positive pairs.
A masked language modeling-based word replacement method has been proposed to alleviate this problem~\citep{chuang2022diffcse}.
They train a sentence encoder to predict the replaced words and make the encoder aware of surface-level augmentations.
Some studies adopt momentum contrastive learning to generate positive samples inspired by unsupervised visual representation learning~\citep{zhang2021bootstrapped, wu2021esimcse}.
Prompting~\citep{jiang2022promptbert, jiang2022deep} is another direction that is capable of generating positive pairs.
Recently, a negative sampling method for data augmentation has been proposed~\citep{wang2022sncse}.
This approach takes the negation of an input sentence and uses this soft-negative sample in a contrastive learning framework.
Compared to previous approaches focused on generating better positive and negative pairs, our work uses nearest-neighbor sentences to predict better semantic vectors of sentences, a novel approach that previous approaches have yet to cover.
Related but different from our work, Trans-Encoder proposes the self-distillation method that gets supervision signals from itself~\citep{liu2022transencoder}.
Trans-Encoder solves a slightly different problem from ours. 
They aim to solve an unsupervised sentence pair modeling problem, not unsupervised sentence embedding; although this work does not require any human-annotated similarity scores of sentence pairs for training, they need the sentence pairs of the STS datasets, which are not allowed to be used for training in unsupervised sentence representation learning.}
\section{Method}
\yeon{Leveraging the $k$-nearest-neighbor sentences helps a sentence encoder to approximate a more accurate semantic meaning of a sentence.
For instance, when two input sentences have more common neighbors than other sentences, it is likely that they are semantically similar; we have provided an example in Figure \ref{fig:intro_example}.
We extend this idea to leverage the entire sentences in the corpus, not just the neighbor sentences.
Our unsupervised sentence encoder, RankEncoder, computes a rank vector for a given sentence.
The rank vector is a list of ranks of all sentences in the corpus computed by their similarity scores to the input; for a given corpus with $n$ number of sentences, a rank vector is an $n$-dimensional vector, in which $i$'th element represents the rank of $i$'th sentence in the corpus.
Thus, when two input sentences have common neighbor sentences, their rank vectors are similar.
We found that rank vectors capture more accurate semantic similarity than previous unsupervised sentence encoders.
Since rank vectors predict better semantic similarity scores between sentences, we use these scores for training another sentence encoder to further increase its STS performance.
We provide the overall illustration of RankEncoder in Figure \ref{fig:method}.}
\subsection{Contrastive Learning for Base Encoder}
The first step of our framework is to learn a base sentence encoder $E_1$ via the standard contrastive learning approach~\citep{pmlr-v119-chen20j}. 
Given an input sentence $x_i$, we first create a positive example $x_i^+$ which is semantically similar to $x_i$~\citep{gao-etal-2021-simcse,chuang2022diffcse}; we apply each data augmentation method used by existing unsupervised sentence representation learning studies~\citep{gao-etal-2021-simcse, jiang2022promptbert, wang2022sncse} and verify that our approach works in all cases.
Then, a text encoder, e.g., BERT~\citep{devlin-etal-2019-bert} and RoBERTa~\citep{liu2019roberta}, predicts their sentence vectors, $\vec{\vv}_i$ and $\vec{\vv}^+_i$.
Given a batch of $m$ sentences $\{x_i\}_{i=1}^m$, the contrastive training objective for the sentence $x_i$ with in-batch negative examples is as follows:
\begin{equation}
    l_{i} = -\log \frac{e^{\text{cos}(\vec{\vv_i},\vec{\vv}_i^+)/\tau}}{\sum_{j=1}^{m}e^{\text{cos}(\vec{\vv}_i,\vec{\vv}_j^+ )/\tau}},
    \label{eq:cl_loss}
\end{equation}
where $\text{cos}(\cdot)$ is the cosine similarity function and $\tau$ is the temperature hyperparameter. 
We then get the overall contrastive loss for the whole batch by summing over all the sentences; $l_{cl} = \sum_{i=1}^m l_{i}$. Note that the training objective $l_{cl}$ can be further enhanced by adding other relevant losses~\citep{chuang2022diffcse}, transforming the input sentences~\citep{jiang2022promptbert, gao-etal-2021-simcse}, or modifying the standard contrastive loss~\citep{zhou-etal-2022-debiased}. 
For simplicity, we use $l_{cl}$ to represent all the variants of contrastive learning loss in this paper. 
By optimizing $l_{cl}$, we obtain a coarse-grained sentence encoder $E_1$ for the following steps.

\subsection{RankEncoder}\label{sec:rank_encoder}
RankEncoder computes the orders of sentences in the corpus with their similarity scores to the input sentence.
For a given corpus with $n$ sentences, $\mathcal{C}=[x_1,...,x_n]$, and a given base encoder, $E_1$, RankEncoder computes the vector representation of each sentence in the corpus, $\mathcal{V}=[\vec{\vv}_1,...,\vec{\vv}_n]$, with $E_1$.
Then computes the rank vector of an input sentence, $x$, by their orders as follows:
\begin{equation}\label{eq:rank_vector}
    \text{RankEncoder}_{E_1}(x, \mathcal{V}) = g(<r_1, r_2, ..., r_n>),
\end{equation}
where $r_i$ is the order of sentence $x_i$. 
We use cosine similarity scores between $\mathcal{V}$ and the vector representation of $x$, $E_1(x)$. 
The function $g$ is a normalization function defined as follows:
\begin{equation}
    g(\vec{\bf{r}}) = \frac{\vec{\bf{r}} - \frac{1}{n}\sum_{i=1}^n r_i\cdot \vec{\bm{1}}}{\sqrt{n} \times \sigma([r_i]_{i=1}^n)},
\end{equation}
where $\sigma$ is the standard deviation of the input values, $\vec{\bm{1}}$ is a vector of ones of size $n$.
By applying this function to rank vectors, the inner product of two normalized rank vectors becomes equivalent to Spearman’s rank correlation, and the similarity is scaled between -1 and 1.
We describe the connection between normalization function $g$ and Spearman’s rank correlation in Appendix \ref{sec:appendix_rho}.

\begin{figure}[ht]
    \centering
    \includegraphics[width=0.95\linewidth]{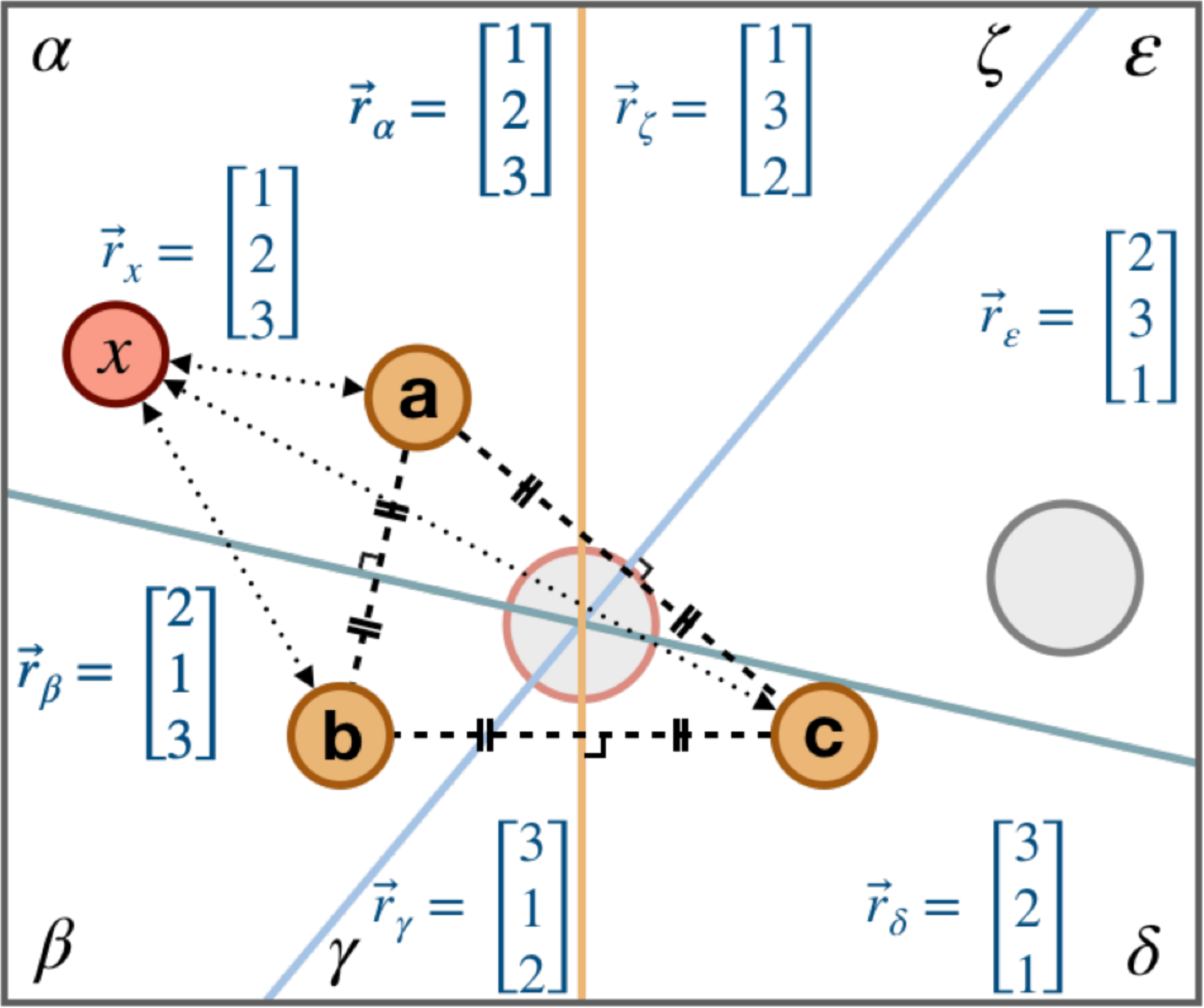}
    \caption{The rank vector space of RankEncoder with corpus $\mathcal{C}=\{a, b, c\}$. 
    The sentence vectors are computed by the base encoder, $E$. 
    Each element of $\vec{\bf{r}}$ corresponds to the rank of each sentence; the first element in the vector is the rank of sentence $a$.
    Each line represents the boundary that two rank variables are converted. 
    For instance, all vectors on the left of the yellow line are closer to $b$ than $c$.
    Sentence vectors in the same area have the same rank vector; the rank vector of sentence $x$ is the same as $\vec{r}_{\alpha}$ as it is in the $\alpha$ area.}
    \label{fig:rank_vector_space}
    \vspace{-1em}
\end{figure}
\subsection{Semantic Vector Space of Rank Vectors}\label{sec:method_vec_space_of_rank_encoder}
\yeon{The similarity between rank vectors is affected mainly by their neighbor sentences, even though we use the entire sentences in a given corpus.
Figure \ref{fig:rank_vector_space} shows the example of RankEncoder’s semantic space when the corpus has three sentences. 
Each solid line represents the boundary that two rank variables are converted. 
For instance, the yellow line is the boundary that reverses the orders of sentences $b$ and $c$; all the vectors located in the left part of the yellow line are closer to sentence $b$ than $c$. 
Since we have three sentences in this corpus, we get six rank vectors, and all vectors in each region have the same rank vector.
In this figure, we see that the vectors in the red area are more affected by sentences $a$, $b$, and $c$ than vectors in the grey area.
For a given sentence, if its sentence representation lies in the central area, i.e., the red area, then its corresponding rank vector can be easily changed by a small modification of its sentence vector.
For vectors having a larger distance from these sentences, e.g., the vectors in the gray area, the corresponding rank vectors are much less sensitive to modification of the input's sentence vector.
This pattern also holds when we increase the size of the corpus as well; we demonstrate this in Section \ref{sec:exp_vector_space}.}

\subsection{Model Training}
We use similarities predicted by rank vectors to train another sentence encoder, $E_2$\footnote{\revisionyeon{It is also possible to train $E_1$ continuously with the rank vector similarities. However, this approach yields slightly lower performance than training another sentence encoder, $E_2$.}}; rank vectors capture a better semantic similarity than their vector representation computed by base encoder $E_1$.
For a given unsupervised sentence encoder $E_1$ and corpus $\mathcal{C}$, we compute similarity scores of all sentence pairs in a batch with their rank vectors computed by $E_1$.
The similarity scores are computed by the inner product of these rank vectors.
Then, we define the loss function as the mean square error of RankEncoder’s similarity scores as follows:
\begin{equation}
    l_{\text{r}} = \frac{1}{m^2}\sum_{i=1}^{m}\sum_{j=1}^m{\Big(\vec{\bf{u}}_i^{\intercal}\vec{\bf{u}}_j-\text{cos}\big(E_2(x_i), E_2(x_j)\big)\Big)^2},
\end{equation}
where $\{x_i\}_{i=1}^m$ are the sentences in the batch, $E_2$ is the sentence encoder in training, $\vec{\bf{u}}_i$ is a rank vector of $x_i$ computed by $\text{RankEncoder}_{E_1}$, and $\text{cos}(\cdot)$ is the cosine similarity function.
Then, we combine the RankEncoder loss, $l_r$, with the standard contrastive loss, $l_{\text{cl}}$, in the form of the hinge loss as follows:
\begin{equation}\label{eq:loss}
    l_{\text{total}} = max(\lambda_{\text{train}} \times l_r, l_{\text{cl}}),
\end{equation}
where $\lambda_{\text{train}}$ is a weight parameter.

\subsection{Sentence Pair Filtering}\label{sec:method_sentence_pair_filtering}
Previous unsupervised sentence encoders randomly sample sentences to construct a batch, and randomly sampled sentence pairs are mostly dissimilar pairs.
This causes sentence encoders to learn mostly on dissimilar pairs, which is less important than similar sentence pairs.
To alleviate this problem, we filter dissimilar sentence pairs with a similarity under a certain threshold\footnote{\revisionyeon{This method increases RankEncoder's STS performance by 0.17\% Spearman correlation, resulting in a performance of 80.07 Spearman correlation shown in Table \ref{tab:sts}.}}.
Also, it is unlikely that randomly sampled sentence pairs have the same semantic meaning. 
We regard sentence pairs with high similarity as noisy samples and filter these pairs with a certain threshold.
The final RankEncoder loss function with sentence pair filtering is as follows:
\begin{equation}
    \begin{split}
        l_r = \sum_{i=1}^{m}&\sum_{j=1}^m\Big\{\frac{\mathbbm{1}[\tau_l \leq \vec{\bf{u}}_i^{\intercal}\vec{\bf{u}}_j \leq \tau_u]}{{\sum_{p=1}^{m}\sum_{q=1}^m\mathbbm{1}[\tau_l \leq \vec{\bf{u}}_p^{\intercal}\vec{\bf{u}}_q \leq \tau_u]}}\\
        &\times\Big(\vec{\bf{u}}_i^{\intercal}\vec{\bf{u}}_j-\text{cos}\big(E_2(x_i), E_2(x_j)\big)\Big)^2\Big\},
    \end{split}
\end{equation}
where $\tau_l$ and $\tau_u$ are the thresholding parameters, and $\mathbbm{1}$ is the indicator function that returns $1$ when the condition is true and returns $0$ otherwise.

\subsection{Inference}\label{sec:method_inference}
We can further utilize RankEncoder in inference stage. 
Given a sentence pair $(x_i, x_j)$, we compute the similarity between the two sentences as follows:
\begin{equation}\label{eq:inference}
    \begin{split}
        \text{sim}&(x_i, x_j) = \lambda_{\text{inf}}\cdot \vec{\bf{z}}_i^{\intercal}\vec{\bf{z}}_j \\
        &+ (1-\lambda_{\text{inf}})\cdot \text{cos}\big(E_2(x_i), E_2(x_j)\big),
    \end{split}
\end{equation}
where $E_2$ is a sentence encoder trained by Eq \ref{eq:loss}, $\lambda_{\text{inf}}$ is a weight parameter, and $\vec{\bf{z}}_i$ and $\vec{\bf{z}}_j$ are the rank vectors of $x_i$ and $x_j$ computed by $\text{RankEncoder}_{E_2}$.
\section{Experimental Setup}
\subsection{Base Encoder $E_1$ \& Corpus $\mathcal{C}$}
RankEncoder computes rank vectors using corpus $\mathcal{C}$ and base encoder $E_1$.
We use 100,000 sentences sampled from Wikipedia\footnote{\revisionyeon{We use Wikipedia sentences since they are generally effective on STS datasets. However, for optimal results, it is best to use a corpus that closely aligns with the specific domain of the inputs.}} as the corpus ($\mathcal{C}$)\footnote{\revisionyeon{The performance of RankEncoder increases as the number of sentences in the corpus increases, and the performance converges at a size of 10,000. There is a slight but negligible improvement beyond this point. We sample more than 10,000 sentences to push the performance boundary as much as possible.}}, and we use the following unsupervised sentence encoders for $E_1$, SimCSE~\citep{gao-etal-2021-simcse}, PromptBERT~\citep{jiang2022promptbert}, and SNCSE~\citep{wang2022sncse}.
SimCSE is a standard unsupervised sentence encoder that uses a standard contrastive learning loss with the simple data augmentation method.
We use SimCSE as it is effective to show the efficacy of RankEncoder.
We use PromptBERT and SNCSE, the state-of-the-art unsupervised sentence encoders, to verify whether RankEncoder is effective on more complex models.

\begin{table*}[t]
\centering
%\resizebox{\columnwidth}{60}{
\begin{tabular}{@{}lcccccccc@{}}
\toprule
Model                                              & {STS12}          & {STS13}          & {STS14}          & {STS15}          & {STS16}          & {STS-B}          & {SICK-R}         & {AVG}            \\ \midrule
ConSERT{\tiny~\citep{yan2021consert}}         & 64.64          & 78.49          & 69.07          & 79.72          & 75.95          & 73.97          & 67.31          & 72.74          \\
SimCSE{\tiny~\citep{gao-etal-2021-simcse}}    & 68.40          & 82.41          & 74.38          & 80.91          & 78.56          & 76.85          & 72.23          & 76.25          \\
DCLR{\tiny~\citep{zhou-etal-2022-debiased}}  & 70.81          & 83.73          & 75.11          & 82.56          & 78.44          & 78.31          & 71.59          & 77.22          \\
ESimCSE{\tiny~\citep{wu2021esimcse}}          & 73.40          & 83.27          & 77.25          & 82.66          & 78.81          & 80.17          & 72.30          & 78.27          \\
DiffCSE{\tiny~\citep{chuang2022diffcse}}      & 72.28          & 84.43          & 76.47          & 83.90          & 80.54          & 80.59          & 71.23          & 78.49          \\
PromptBERT{\tiny~\citep{jiang2022promptbert}} & 71.56          & 84.58          & 76.98          & \textbf{84.47} & \textbf{80.60} & \textbf{81.60} & 69.87          & 78.54          \\
SNCSE{\tiny~\citep{wang2022sncse}}            & 70.67          & 84.79          & 76.99          & 83.69          & 80.51          & 81.35          & 74.77          & 78.97          \\
\rowcolor[HTML]{EFEFEF} RankEncoder                & \textbf{74.88} & \textbf{85.59} & \textbf{78.61} & 83.50          & 80.56          & 81.55          & \textbf{75.78} & \textbf{80.07} \\ \bottomrule
\end{tabular}
%}
\caption{Semantic textual similarity performance of RankEncoder and baselines in an unsupervised setting. Following previous sentence embedding studies, we measure the Spearman's rank correlation between the human annotated scores and the model's predictions. The results of the baselines are from the original paper. RankEncoder uses SNCSE as base encoder $E_1$.}
\label{tab:sts}
\vspace{-1em}
\end{table*}

\subsection{Datasets \& Evaluation Metric}
We evaluate RankEncoder on seven semantic textual similarity benchmark datasets: STS2012-2016~\citep{agirre2012semeval, agirre2013sem, agirre-etal-2014-semeval, agirre2015semeval, agirre-etal-2016-semeval}, STS-B~\citep{cer-etal-2017-semeval}, and SICK-Relatedness~\citep{marelli-etal-2014-sick}. Each dataset consists of sentence pairs with human-annotated similarity scores. 
For each sentence pair, sentence encoders predict the similarity, and we measure the Spearman’s rank correlation between the predicted similarities and the human-annotated similarities.

\subsection{Training Details \& Hyper-Parameter Settings}
We train RankEncoder on $10^6$ sentences from Wikipedia, following existing unsupervised sentence embedding studies. 
\yeon{We use two NVIDIA V100 GPUs for training. The running time for training RankEncoder is approximately 1.5 hours, which takes an hour more than the training time of SimCSE, and its inference takes slightly more time, about 1.8\% more than SimCSE based on BERT-base.}
\revisionyeon{We provide the details in Appendix \ref{appendix:computation_cost}}.
We find the best hyperparameter setting on the development sets of the STS-B and SICKRelatedness datasets.
We set $\lambda_{\text{train}}=0.05$, $\lambda_{\text{inf}}=0.1$, $\tau_l=0.5$, and $\tau_u=0.8$.
We provide more analysis on the hyper-parameter, $\lambda_{\text{train}}$, in Appendix \ref{sec:lmb_analysis}.
For other hyperparameters, we follow the base encoder’s setting provided by the authors of each base encoder, $E_1$.
\section{Results and Discussions}
\begin{figure*}[ht]
    \begin{center}
        \begin{subfigure}{.32\textwidth}
            \centering
            \includegraphics[width=1.\linewidth]{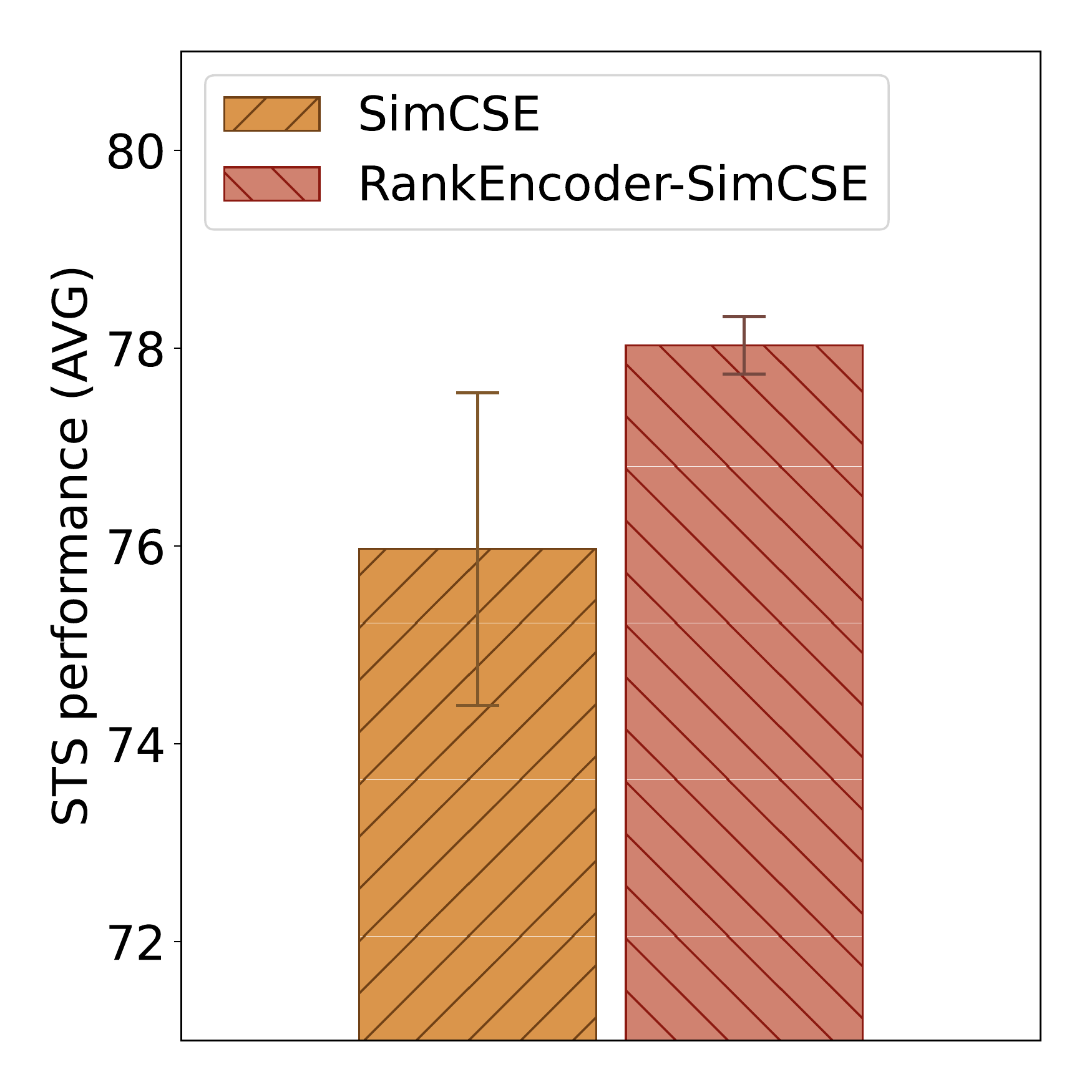}
            \label{fig:inc_simcse}
        \end{subfigure}
        \begin{subfigure}{.32\textwidth}
            \centering
            \includegraphics[width=1.\linewidth]{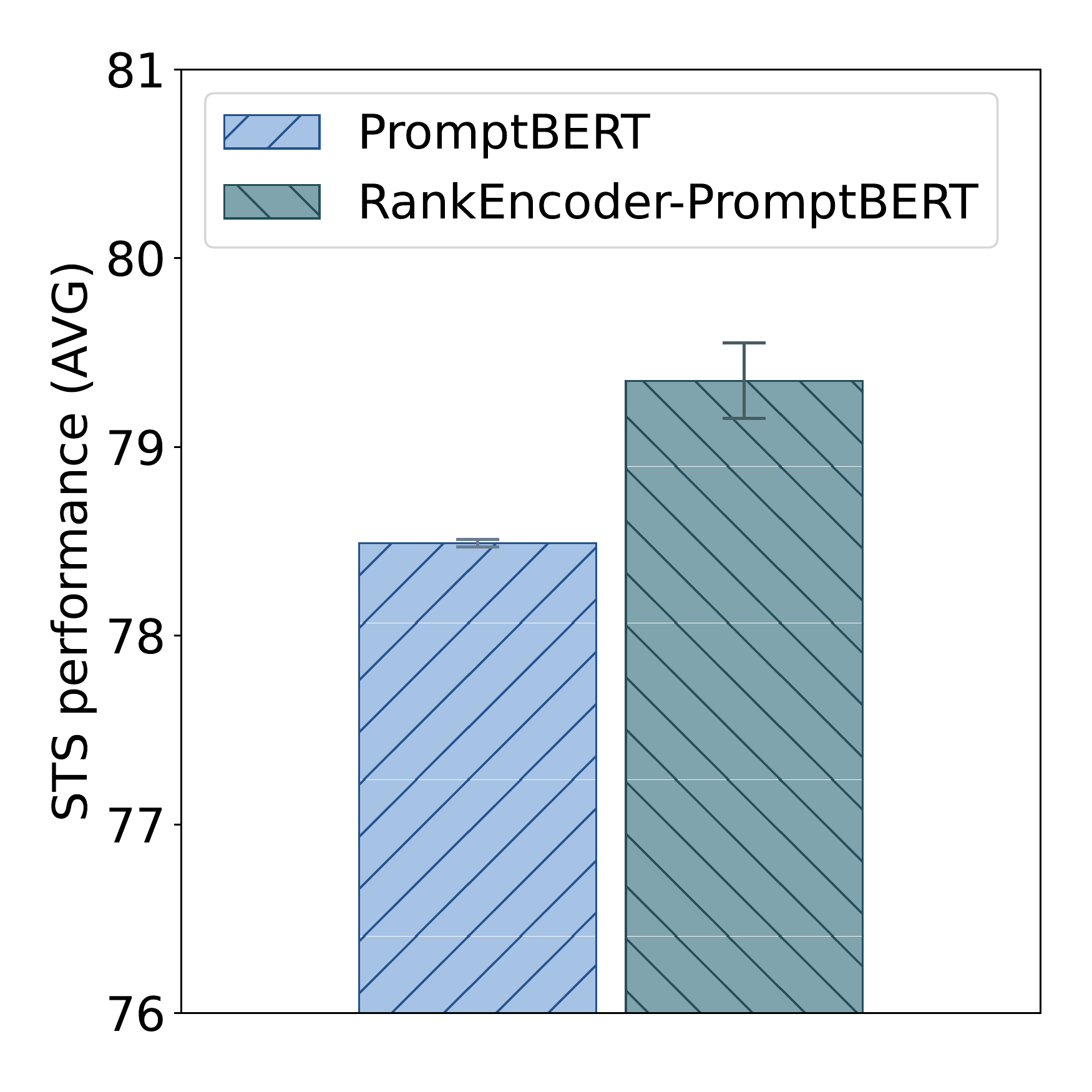}
            \label{fig:inc_promptbert}
        \end{subfigure}
        \begin{subfigure}{.32\textwidth}
            \centering
            \includegraphics[width=1.\linewidth]{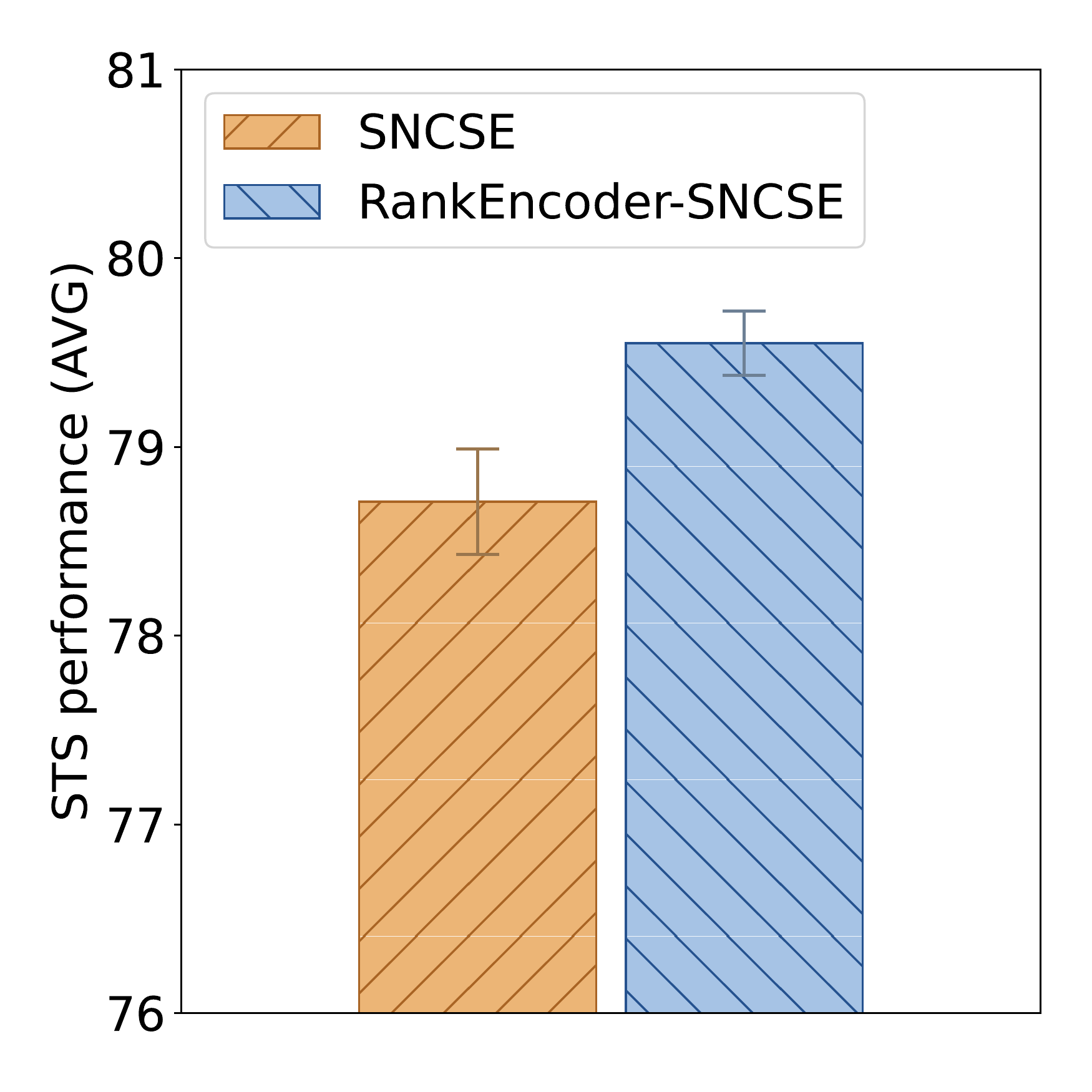}
            \label{fig:inc_sncse}
        \end{subfigure}
    \vspace{-1em}
    \caption{STS performance of three unsupervised sentence encoders and RankEncoder. We report the mean performance and standard deviation of three separate trials with different random seeds. RankEncoder brings improvement on all base encoders.
    This result implies that our approach is generally applicable to other unsupervised sentence embedding approaches.
    }
    \label{fig:inc}
    \end{center}
\end{figure*}

In this section, we demonstrate that 1) RankEncoder is effective for capturing the semantic similarity scores of similar sentences, 2) RankEncoder is universally applicable to existing unsupervised sentence encoders, and 3) RankEncoder achieves state-of-the-art semantic textual similarity (STS) performance.
We describe the detailed experimental results in the following sections.

\subsection{Semantic Textual Similarity Performance}
We apply RankEncoder to an existing unsupervised sentence encoder and achieve state-of-the-art STS performance.
We use SNCSE~\citep{wang2022sncse} fine-tuned on BERT-base~\citep{devlin-etal-2019-bert} as the base encoder, $E_1$.
Table \ref{tab:sts} shows the STS performance of RankEncoder and unsupervised sentence encoders on seven STS datasets and their average performance (AVG).
RankEncoder increases the AVG performance of SNCSE by 1.1 and achieves state-of-the-art STS performance.

RankEncoder brings a significant performance gain on STS12, STS13, STS14, and SICK-R, but a comparably small improvement on STS16 and STS-B.
We conjecture that this is because RankEncoder is specifically effective on similar sentence pairs.
The STS12, 13, 14, and SICK-R datasets contain similar sentence pairs more than dissimilar pairs; we show the similarity distribution of each dataset in Appendix \ref{sec:appendix_sim_dist}.
This pattern is aligned with the performance gain on each STS dataset in Table \ref{tab:sts}.

\begin{figure*}[t]
    \begin{center}
        \begin{subfigure}{.325\textwidth}
            \centering
            \includegraphics[width=1.\linewidth]{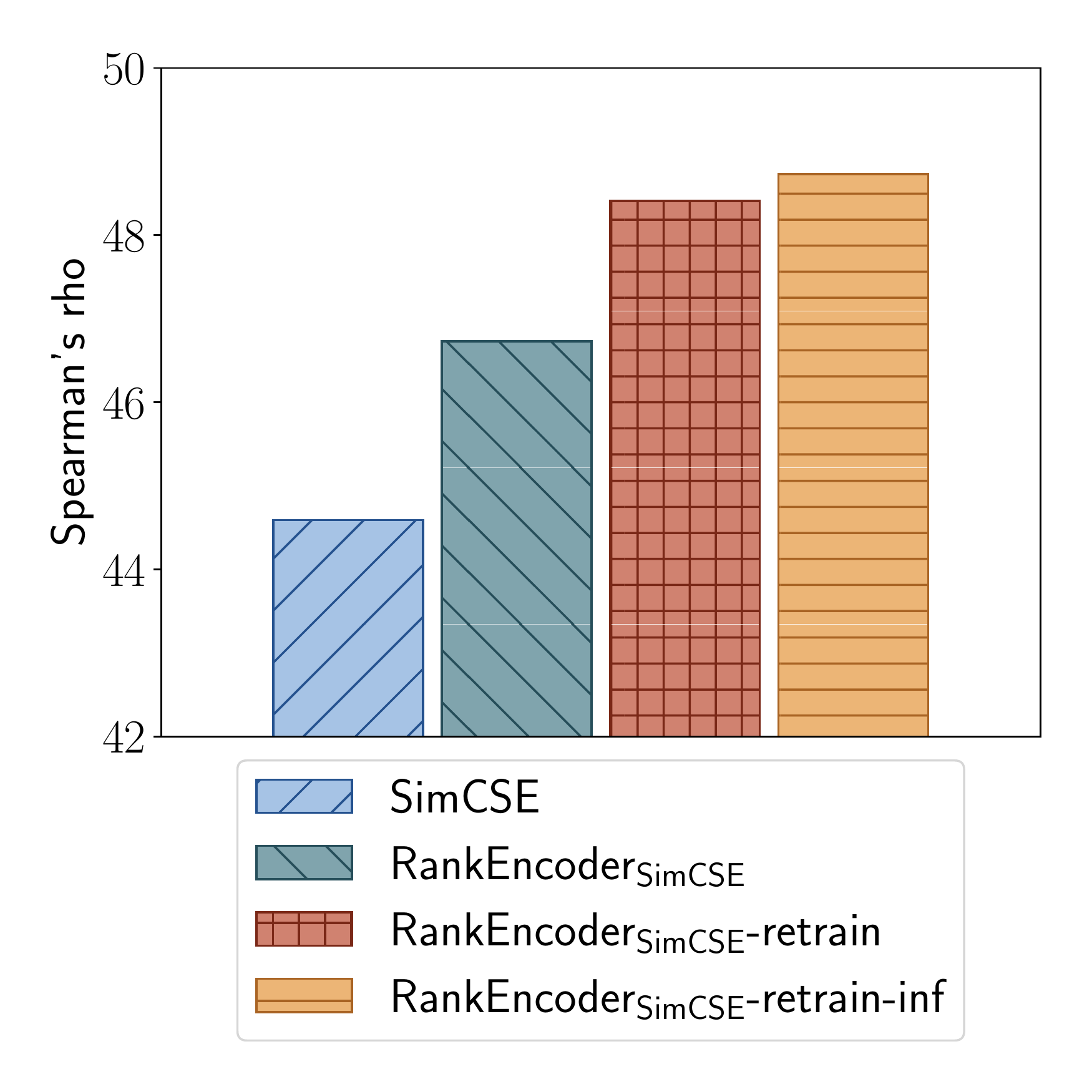}
            \label{fig:perf_similar_pairs_simcse}
        \end{subfigure}
        \begin{subfigure}{.325\textwidth}
            \centering
            \includegraphics[width=1.\linewidth]{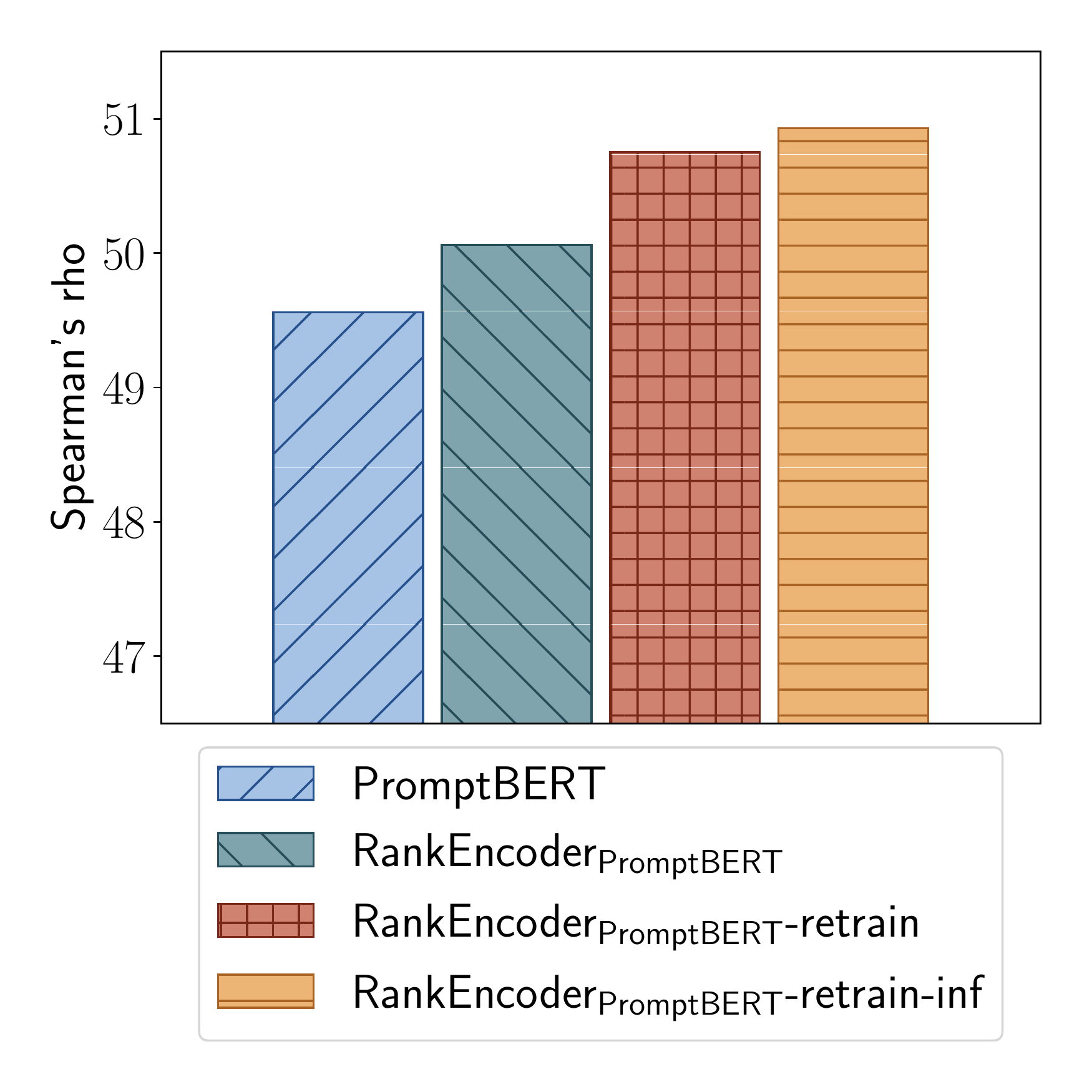}
            \label{fig:perf_similar_pairs_promptbert}
        \end{subfigure}
        \begin{subfigure}{.325\textwidth}
            \centering
            \includegraphics[width=1.\linewidth]{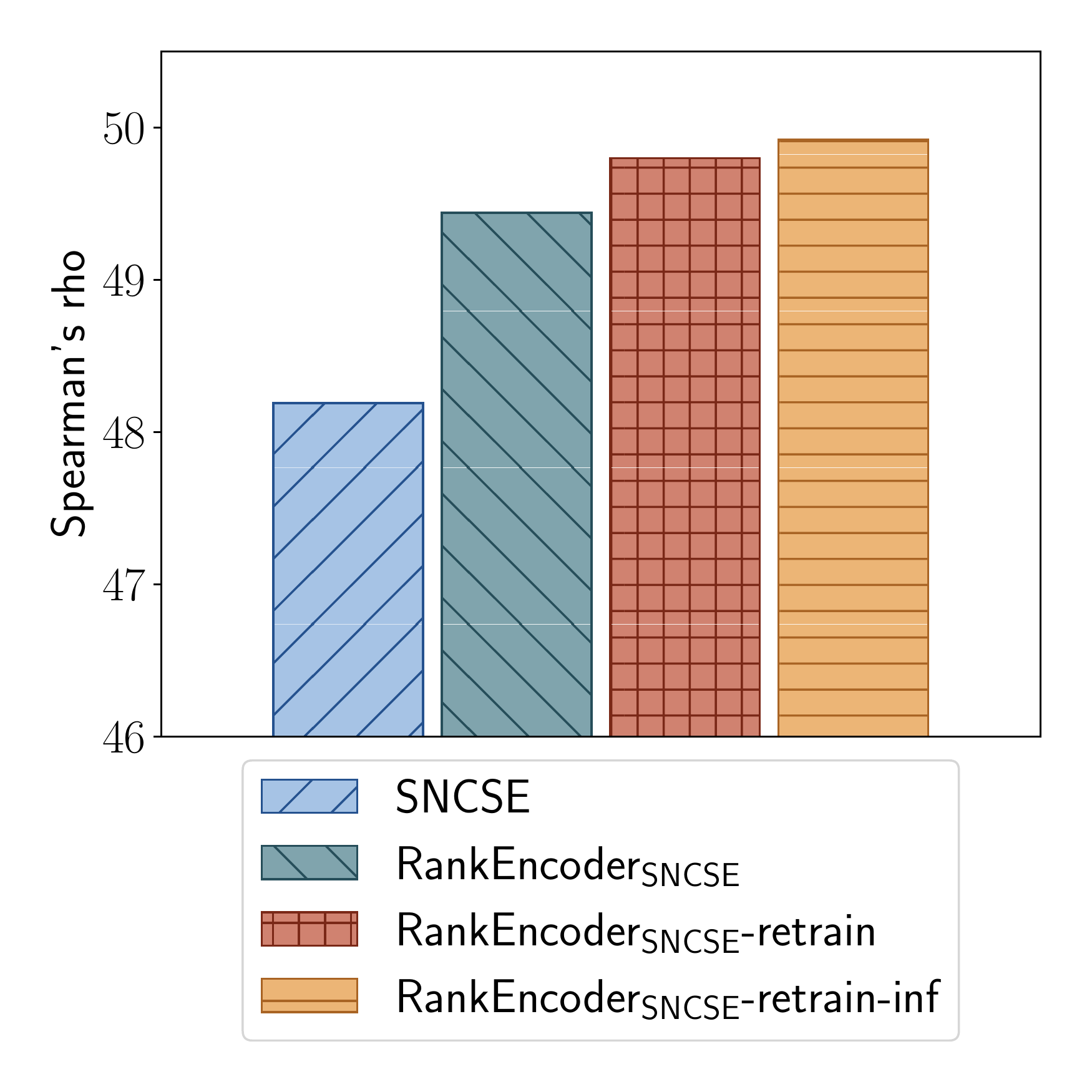}
            \label{fig:perf_similar_pairs_sncse}
        \end{subfigure}
    \vspace{-1em}
    \caption{
    STS performance of three unsupervised sentence encoders and RankEncoder on sentence pairs with high similarity scores; we select sentence pairs with a similarity between 0.67 and 1.0 (0.0-1.0 scale) in the STS-B dataset.
    We ablate the two components of RankEncoder, re-training (Eq. \ref{eq:loss}) and inference (Eq. \ref{eq:inference}).
    }
    \label{fig:perf_similar_pairs}
    \end{center}
    \vspace{-1.5em}
\end{figure*}

\begin{figure}[t]
    \begin{center}
    \includegraphics[width=0.98\linewidth]{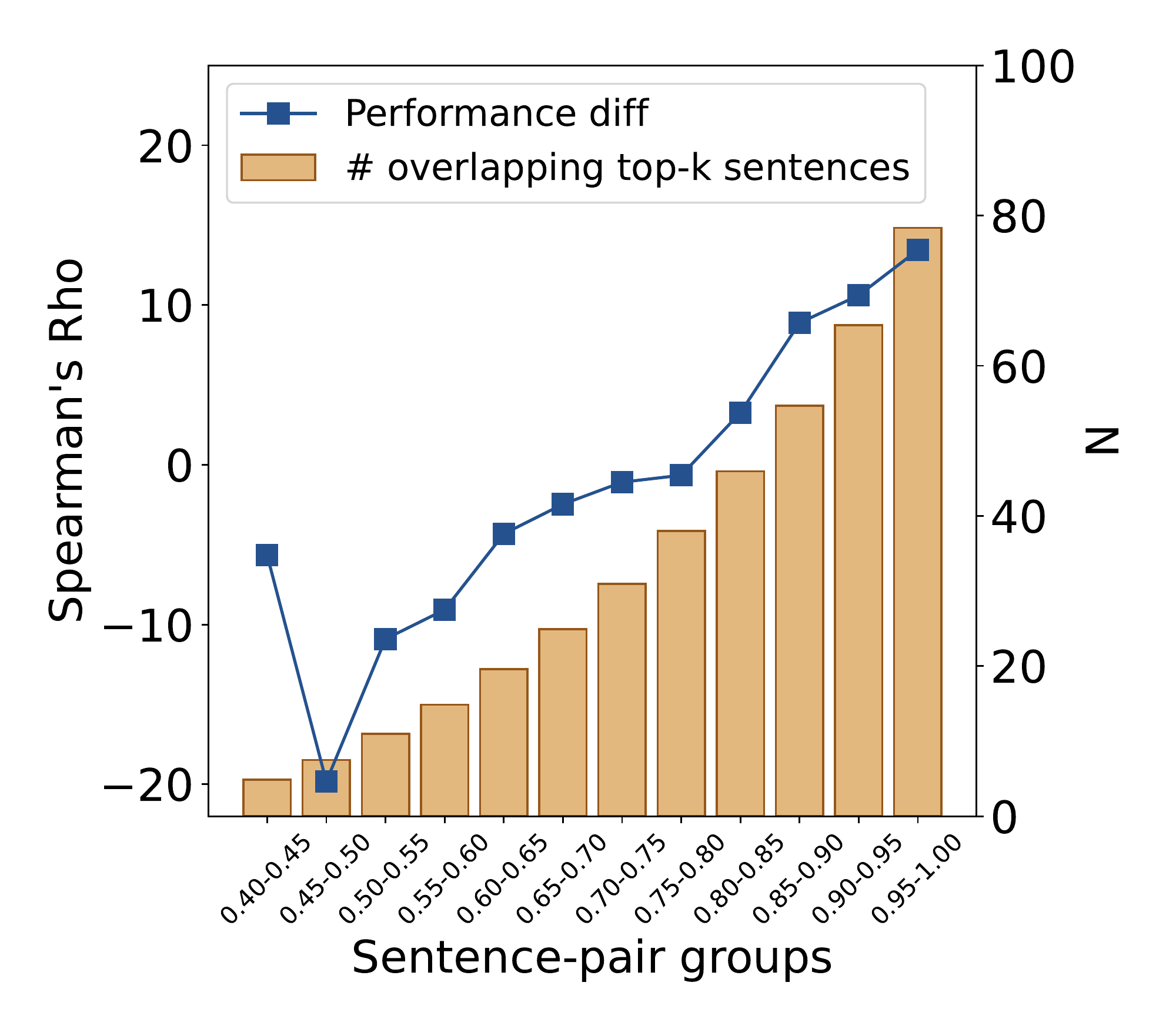}
    \caption{The performance difference between SimCSE and RankEncoder (blue line) and the number of overlapping neighbor sentences (yellow bar) on each sentence pair group.
    We group sentence pairs in the STS-B dataset based on the cosine similarity of their vector representations computed by SimCSE.
    The number of overlapping neighbor sentences is the number of sentences in the intersection of each sentence's top 100 neighbor sentences.
    }
    \label{fig:perf_overlap}
    \end{center}
    \vspace{-1em}
\end{figure}

\subsection{Universality of RankEncoder}\label{sec:univ_rankencoder}
RankEncoder applies to any unsupervised sentence encoders. 
We apply RankEncoder to SimCSE~\citep{gao-etal-2021-simcse}, PromptBERT~\citep{jiang2022promptbert}, and SNCSE~\citep{wang2022sncse}. 
SimCSE represents the vanilla contrastive learning-based sentence encoder, and PromptBERT and SNCSE represent the state-of-the-art unsupervised sentence encoders.
We evaluate each encoder's average performance (AVG) on seven STS datasets.
We train each encoder in three separate trials and report the mean and the standard deviation of the AVG performances in Figure \ref{fig:inc}; the error bar shows the standard deviation. 
This figure shows that RankEncoder increases the average STS performance on each unsupervised sentence encoder; the improvements on SimCSE, PromptBERT, and SNCSE are 2.1, 0.9, and 0.9, respectively.
We report detailed experimental results in Appendix \ref{sec:appendix_universality}.
This result implies that RankEncoder is a universal method that applies to any unsupervised sentence encoder.

\subsection{Overlapping Neighbor Sentences}\label{sec:exp_perf_overlap}
In Section \ref{sec:method_vec_space_of_rank_encoder}, we conjecture that the RankEncoder is specifically effective for similar sentence pairs as they have more overlapping neighbor sentences, which are used to approximate their semantic similarity.
To support this supposition, we show the relation between the performance gain caused by RankEncoder and the number of overlapping neighbor sentences of the input sentences.
We group sentence pairs in the STS-B dataset by cosine similarity scores of their sentence vectors, then compare the STS performance of SimCSE and RankEncoder (Eq. \ref{eq:rank_vector} without re-training) on each group; we use SimCSE as the base encoder, $E_1$.
We also report the average number of overlapping neighbor sentences of each group; for each sentence pair, we count the number of sentences in the intersection of their top 100 nearest neighbor sentences and take the average.
Figure \ref{fig:perf_overlap} shows one expected result of our supposition; the performance gain correlates with the number of overlapping neighbor sentences.

\subsection{Performance on similar sentence pairs}
\revisionyeon{It is more important to accurately predict the similarities between similar texts than those between dissimilar ones. This is because many NLP downstream tasks, e.g., retrieval and reranking, aim to find the most relevant/similar text (or texts) from candidate texts, and we can easily filter out dissimilar texts with simple approaches such as a lexical retriever; we only need rough similarity scores to identify the dissimilar texts.}
\revisionyeon{In this section, we demonstrate the efficacy of our approach on similar sentence pairs.}
We divide sentence pairs in the STS-B dataset into three groups by their human-annotated similarity scores and use the group with the highest similarity.
The similarity range of each group is 0.0-0.33 for the dissimilar groups, 0.33-0.67 for the intermediate group, and 0.67-1.0 for the similar group; we normalize the scores to a 0.0-1.0 scale.
Figure \ref{fig:perf_similar_pairs} shows the performance of three unsupervised sentence encoders and the performance gain brought by each component of RankEncoder.
$\text{RankEncoder}_{E}$ is the model with Eq.~\ref{eq:rank_vector} that uses $E$ as the base encoder.
$\text{RankEncoder}_{E}\text{-retrain}$ is the model with re-training (Eq.~\ref{eq:loss}).
$\text{RankEncoder}_{E}\text{-retrain}\text{-inf}$ is the model with re-training and weighted inference (Eq.~\ref{eq:inference}).
From the comparison between $E$ and $\text{RankEncoder}_{E}$, we verify that rank vectors effectively increase the base encoder's performance on similar sentence pairs.
This improvement is even more significant when using rank vectors for re-training and inference.
We report the detailed results in Appendix \ref{sec:appendix_pef_similar_pairs}.

\begin{figure}[t]
    \centering
    \begin{subfigure}{.235\textwidth}
        \centering
        \includegraphics[width=1.\linewidth]{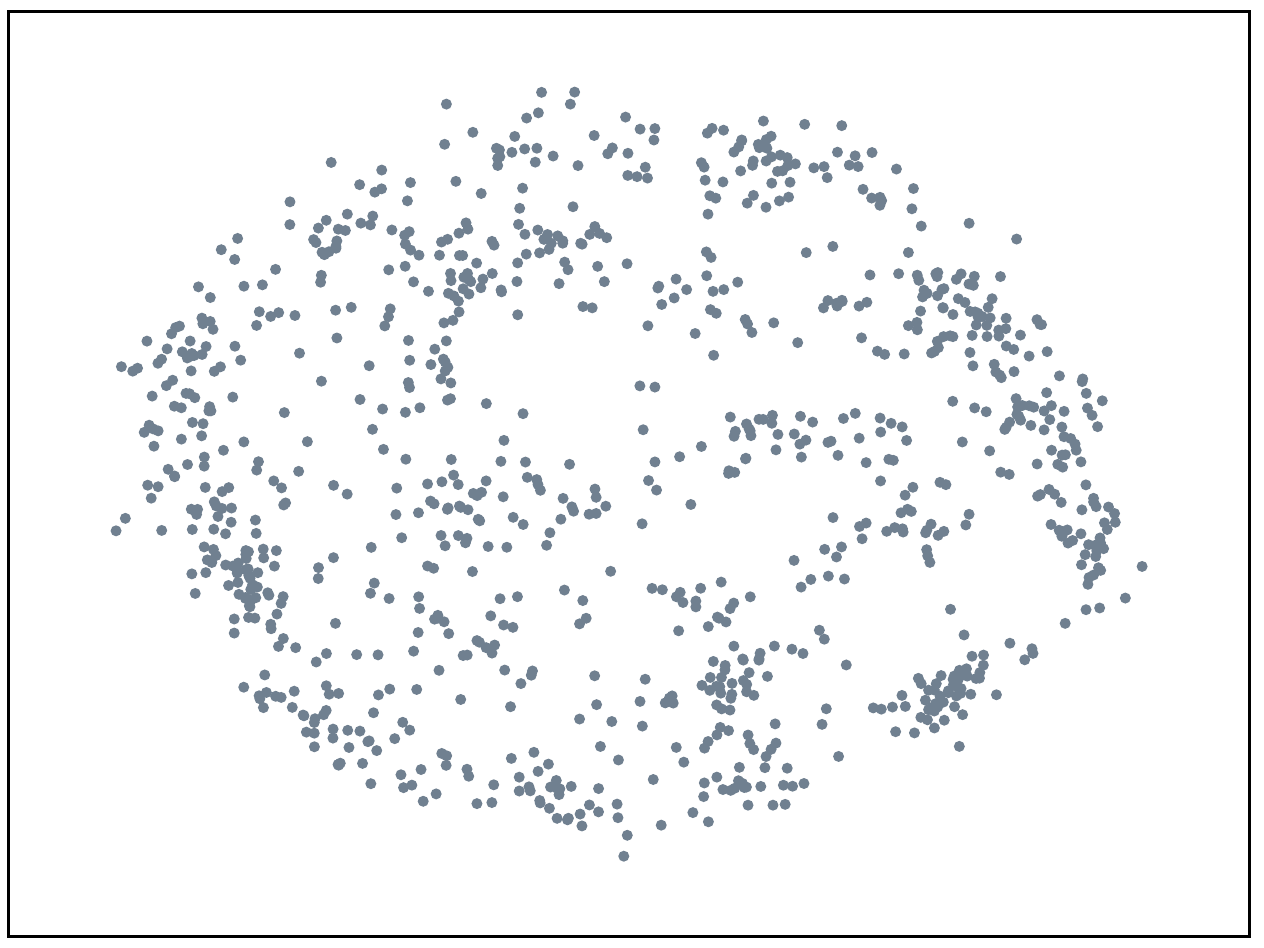}
        \caption{PromptBERT}
        \label{fig:semantic_space_a}
    \end{subfigure}
    \begin{subfigure}{.235\textwidth}
        \centering
        \includegraphics[width=1.\linewidth]{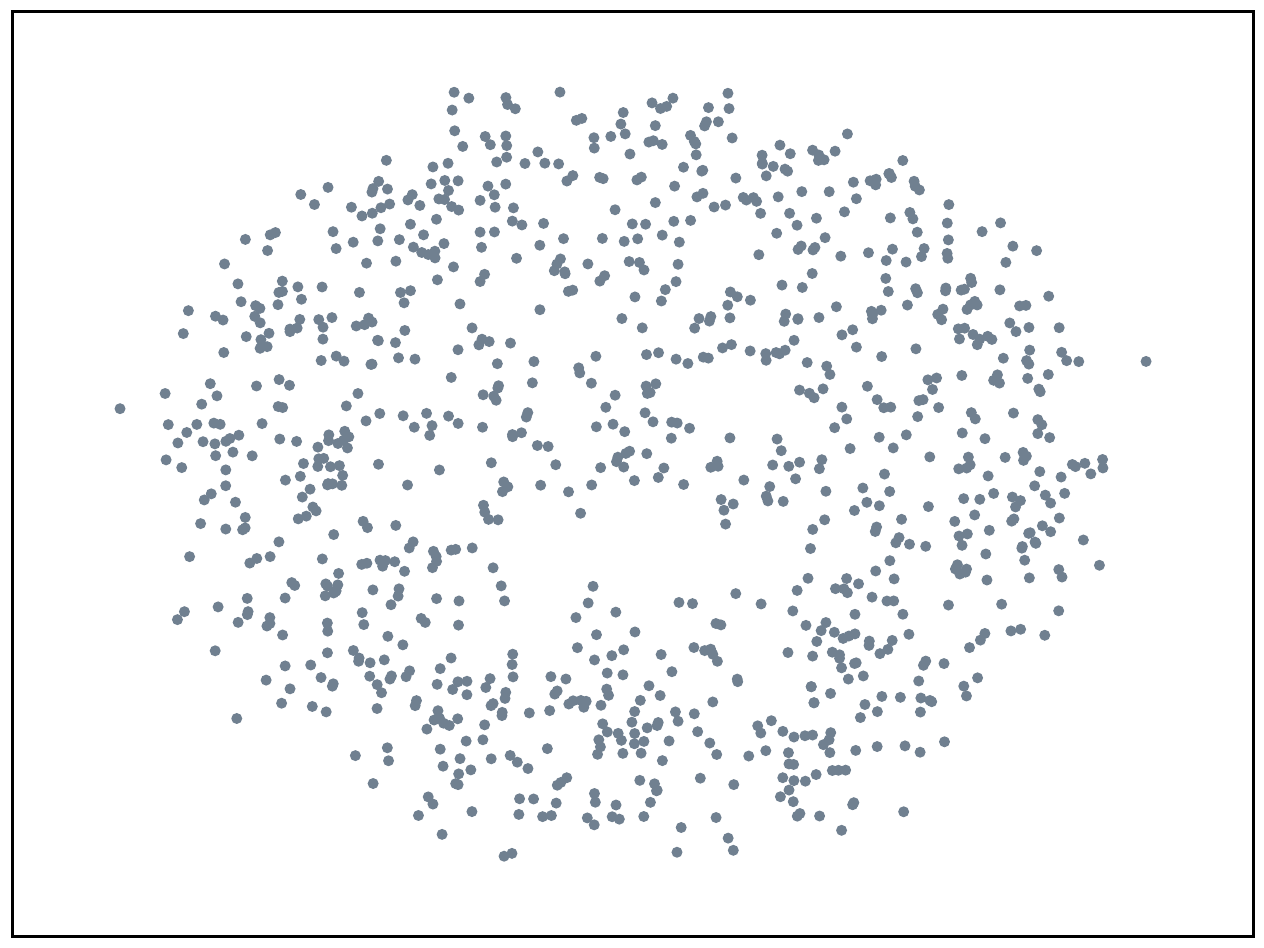}
        \caption{RankEncoder}
        \label{fig:semantic_space_b}
    \end{subfigure}
    \caption{Semantic vector spaces of PromptBERT and RankEncoder.
    We randomly sample 1000 sentences from the STS-B dataset and visualize the vector representations of these sentences (grey dots). We use PromptBERT as the base encoder of RankEncoder. We use the following equation to compute the distances between vectors; $\text{dist}(\vec{\bf{v}}_i, \vec{\bf{v}_j}) = 1-\text{cos}(\vec{\bf{v}}_i, \vec{\bf{v}_j})$. 
    }
    \label{fig:semantic_space}
\end{figure}

\subsection{The Vector Space of RankEncoder}\label{sec:exp_vector_space}
In Section \ref{sec:method_vec_space_of_rank_encoder}, we show that rank vectors become more distinguishable as the number of sharing neighbor sentences increases.
In this section, we demonstrate that this pattern holds for a larger corpus as well.
Figure \ref{fig:semantic_space} shows the vector representations of randomly sampled 1,000 sentences in the STS-B dataset; Figure \ref{fig:semantic_space_a} is the vector space of PromptBERT, and Figure \ref{fig:semantic_space_b} is the rank vector space.
We see that the dense sub-spaces in Figure \ref{fig:semantic_space_a} expand as shown in \ref{fig:semantic_space_b}, and their representations become more distinguishable.

\yeon{Rank vectors improve the uniformity of the semantic vector space with negligible degradation in alignment.
Uniformity and alignment are metrics for measuring the quality of embedding vectors~\citep{gao-etal-2021-simcse, wang2020understanding}.
Uniformity is a measure of the degree of evenness of the embedding vectors.
Alignment is a measure of the degree of closeness of the embedding vectors of positive pairs (e.g., sentence pairs with a similarity score higher than 4.0 in the STS-B dataset).
We show the uniformity and alignment of each base encoder, $E_1$, and their corresponding rank vectors, $\text{RankEncoder}_{E_1}$, in Table \ref{tab:uniformity}.
For each base encoder, their rank vectors largely improve uniformity, which is aligned with the results shown in Figure \ref{fig:semantic_space}.
These results also show that rank vectors bring degradation in alignment.
However, this degradation is relatively negligible compared to the improvement in uniformity.
From these results, we conjecture that the performance improvement on the STS benchmark datasets shown in Figure \ref{fig:inc} is mostly related to the improvement in uniformity rather than alignment.}

\begin{table}[t]
\centering
\begin{tabular}{@{}lcc@{}}
\toprule
             & \multicolumn{1}{l}{Uniformity} & \multicolumn{1}{l}{Alignment} \\ \midrule
SimCSE       & -2.42                          & 0.21                          \\
+RankEncoder & -3.23                          & 0.23                          \\ \midrule
PromptBERT   & -1.49                          & 0.11                          \\
+RankEncoder & -3.31                          & 0.22                          \\ \midrule
SNCSE        & -2.21                          & 0.16                          \\
+RankEncoder & -3.20                          & 0.21                          \\ \bottomrule
\end{tabular}
\caption{Uniformity and alignment of base encoders and RankEncoder. Lower is better.}
\label{tab:uniformity}
\end{table}
\section{Conclusion}
\yeon{In this study, we showed that the semantics of a sentence is also determined by its similar sentence, not just the words within the sentence itself.
We proposed RankEncoder to overcome the limitation of the previous sentence representation learning approaches, which are limited to using only the input sentence.
RankEncoder leverages the distance between the input sentence and the sentences in a corpus to predict its semantic vector.
RankEncoder is universally applicable to any unsupervised sentence encoder, resulting in performance improvement, and we demonstrated this with three unsupervised sentence encoders.
We achieved state-of-the-art semantic textual similarity performance by applying our approach to the previous best sentence encoder.
We also showed that our approach is specifically effective for capturing the semantic similarities of similar sentences.
}
\section{Limitations}
\yeon{This work has been studied on the Wikipedia corpus, following the standard experimental setting used in previous unsupervised sentence representation learning studies.
We expect to see many important findings by investigating sentence representation learning on various corpora in different domains such as Bookcorpus~\citep{Zhu_2015_ICCV} and the C4 corpus~\citep{2019t5}.
}

\section*{Acknowledgements}
This research was supported by the Engineering Research Center Program through the National Research Foundation of Korea (NRF) funded by the Korean Government MSIT (NRF-2018R1A5A1059921)

\bibliography{custom}

\begin{thebibliography}{40}
\expandafter\ifx\csname natexlab\endcsname\relax\def\natexlab#1{#1}\fi

\bibitem[{Agirre et~al.(2015)Agirre, Banea, Cardie, Cer, Diab, Gonzalez-Agirre,
  Guo, Lopez-Gazpio, Maritxalar, Mihalcea et~al.}]{agirre2015semeval}
Eneko Agirre, Carmen Banea, Claire Cardie, Daniel Cer, Mona Diab, Aitor
  Gonzalez-Agirre, Weiwei Guo, Inigo Lopez-Gazpio, Montse Maritxalar, Rada
  Mihalcea, et~al. 2015.
\newblock Semeval-2015 task 2: Semantic textual similarity, english, spanish
  and pilot on interpretability.
\newblock In \emph{SemEval}.

\bibitem[{Agirre et~al.(2014)Agirre, Banea, Cardie, Cer, Diab, Gonzalez-Agirre,
  Guo, Mihalcea, Rigau, and Wiebe}]{agirre-etal-2014-semeval}
Eneko Agirre, Carmen Banea, Claire Cardie, Daniel Cer, Mona Diab, Aitor
  Gonzalez-Agirre, Weiwei Guo, Rada Mihalcea, German Rigau, and Janyce Wiebe.
  2014.
\newblock {S}em{E}val-2014 task 10: Multilingual semantic textual similarity.
\newblock In \emph{{S}em{E}val}.

\bibitem[{Agirre et~al.(2016)Agirre, Banea, Cer, Diab, Gonzalez-Agirre,
  Mihalcea, Rigau, and Wiebe}]{agirre-etal-2016-semeval}
Eneko Agirre, Carmen Banea, Daniel Cer, Mona Diab, Aitor Gonzalez-Agirre, Rada
  Mihalcea, German Rigau, and Janyce Wiebe. 2016.
\newblock {S}em{E}val-2016 task 1: Semantic textual similarity, monolingual and
  cross-lingual evaluation.
\newblock In \emph{{S}em{E}val}.

\bibitem[{Agirre et~al.(2012)Agirre, Cer, Diab, and
  Gonzalez-Agirre}]{agirre2012semeval}
Eneko Agirre, Daniel Cer, Mona Diab, and Aitor Gonzalez-Agirre. 2012.
\newblock Semeval-2012 task 6: A pilot on semantic textual similarity.
\newblock In \emph{SemEval}.

\bibitem[{Agirre et~al.(2013)Agirre, Cer, Diab, Gonzalez-Agirre, and
  Guo}]{agirre2013sem}
Eneko Agirre, Daniel Cer, Mona Diab, Aitor Gonzalez-Agirre, and Weiwei Guo.
  2013.
\newblock * sem 2013 shared task: Semantic textual similarity.
\newblock In \emph{SemEval}.

\bibitem[{Cer et~al.(2017)Cer, Diab, Agirre, Lopez-Gazpio, and
  Specia}]{cer-etal-2017-semeval}
Daniel Cer, Mona Diab, Eneko Agirre, I{\~n}igo Lopez-Gazpio, and Lucia Specia.
  2017.
\newblock {S}em{E}val-2017 task 1: Semantic textual similarity multilingual and
  crosslingual focused evaluation.
\newblock In \emph{{S}em{E}val}.

\bibitem[{Cer et~al.(2018)Cer, Yang, Kong, Hua, Limtiaco, John, Constant,
  Guajardo-Cespedes, Yuan, Tar et~al.}]{cer2018universal}
Daniel Cer, Yinfei Yang, Sheng-yi Kong, Nan Hua, Nicole Limtiaco, Rhomni~St
  John, Noah Constant, Mario Guajardo-Cespedes, Steve Yuan, Chris Tar, et~al.
  2018.
\newblock Universal sentence encoder.
\newblock \emph{arXiv}.

\bibitem[{Chen et~al.(2020)Chen, Kornblith, Norouzi, and
  Hinton}]{pmlr-v119-chen20j}
Ting Chen, Simon Kornblith, Mohammad Norouzi, and Geoffrey Hinton. 2020.
\newblock A simple framework for contrastive learning of visual
  representations.
\newblock In \emph{ICML}.

\bibitem[{Chuang et~al.(2022)Chuang, Dangovski, Luo, Zhang, Chang,
  Solja{\v{c}}i{\'c}, Li, Yih, Kim, and Glass}]{chuang2022diffcse}
Yung-Sung Chuang, Rumen Dangovski, Hongyin Luo, Yang Zhang, Shiyu Chang, Marin
  Solja{\v{c}}i{\'c}, Shang-Wen Li, Wen-tau Yih, Yoon Kim, and James Glass.
  2022.
\newblock Diffcse: Difference-based contrastive learning for sentence
  embeddings.
\newblock \emph{arXiv}.

\bibitem[{Conneau and Kiela(2018)}]{conneau2018senteval}
Alexis Conneau and Douwe Kiela. 2018.
\newblock Senteval: An evaluation toolkit for universal sentence
  representations.
\newblock \emph{arXiv}.

\bibitem[{Conneau et~al.(2017)Conneau, Kiela, Schwenk, Barrault, and
  Bordes}]{conneau2017supervised}
Alexis Conneau, Douwe Kiela, Holger Schwenk, Lo{\"\i}c Barrault, and Antoine
  Bordes. 2017.
\newblock Supervised learning of universal sentence representations from
  natural language inference data.
\newblock In \emph{EMNLP}.

\bibitem[{Devlin et~al.(2019)Devlin, Chang, Lee, and
  Toutanova}]{devlin-etal-2019-bert}
Jacob Devlin, Ming-Wei Chang, Kenton Lee, and Kristina Toutanova. 2019.
\newblock {BERT}: Pre-training of deep bidirectional transformers for language
  understanding.
\newblock In \emph{NAACL}.

\bibitem[{Dolan and Brockett(2005)}]{dolan2005automatically}
William~B Dolan and Chris Brockett. 2005.
\newblock Automatically constructing a corpus of sentential paraphrases.
\newblock In \emph{Proceedings of the Third International Workshop on
  Paraphrasing}.

\bibitem[{Gao et~al.(2021)Gao, Yao, and Chen}]{gao-etal-2021-simcse}
Tianyu Gao, Xingcheng Yao, and Danqi Chen. 2021.
\newblock {S}im{CSE}: Simple contrastive learning of sentence embeddings.
\newblock In \emph{EMNLP}.

\bibitem[{Hill et~al.(2016)Hill, Cho, and Korhonen}]{hill2016learning}
Felix Hill, Kyunghyun Cho, and Anna Korhonen. 2016.
\newblock Learning distributed representations of sentences from unlabelled
  data.
\newblock In \emph{NAACL}.

\bibitem[{Hu and Liu(2004)}]{hu2004mining}
Minqing Hu and Bing Liu. 2004.
\newblock Mining and summarizing customer reviews.
\newblock In \emph{KDD}.

\bibitem[{Izacard et~al.(2021)Izacard, Caron, Hosseini, Riedel, Bojanowski,
  Joulin, and Grave}]{izacard2021unsupervised}
Gautier Izacard, Mathild Caron, Lucas Hosseini, Sebastian Riedel, Piotr
  Bojanowski, Armand Joulin, and Edouard Grave. 2021.
\newblock Unsupervised dense information retrieval with contrastive learning.
\newblock \emph{arXiv}.

\bibitem[{Jiang et~al.(2022)Jiang, Huang, Zhang, Wang, Zhuang, Wei, Huang,
  Zhang, and Zhang}]{jiang2022promptbert}
Ting Jiang, Shaohan Huang, Zihan Zhang, Deqing Wang, Fuzhen Zhuang, Furu Wei,
  Haizhen Huang, Liangjie Zhang, and Qi~Zhang. 2022.
\newblock Promptbert: Improving bert sentence embeddings with prompts.
\newblock \emph{arXiv}.

\bibitem[{Jiang and Wang(2022)}]{jiang2022deep}
Yuxin Jiang and Wei Wang. 2022.
\newblock Deep continuous prompt for contrastive learning of sentence
  embeddings.
\newblock \emph{arXiv}.

\bibitem[{Kim et~al.(2021)Kim, Yoo, and Lee}]{kim2021self}
Taeuk Kim, Kang~Min Yoo, and Sang-goo Lee. 2021.
\newblock Self-guided contrastive learning for bert sentence representations.
\newblock In \emph{ACL}.

\bibitem[{Kiros et~al.(2015)Kiros, Zhu, Salakhutdinov, Zemel, Urtasun,
  Torralba, and Fidler}]{kiros2015skip}
Ryan Kiros, Yukun Zhu, Russ~R Salakhutdinov, Richard Zemel, Raquel Urtasun,
  Antonio Torralba, and Sanja Fidler. 2015.
\newblock Skip-thought vectors.
\newblock \emph{NeurIPS}.

\bibitem[{Liu et~al.(2022)Liu, Jiao, Massiah, Yilmaz, and
  Havrylov}]{liu2022transencoder}
Fangyu Liu, Yunlong Jiao, Jordan Massiah, Emine Yilmaz, and Serhii Havrylov.
  2022.
\newblock Trans-encoder: Unsupervised sentence-pair modelling through self- and
  mutual-distillations.
\newblock In \emph{ICLR}.

\bibitem[{Liu et~al.(2021)Liu, Vuli{\'c}, Korhonen, and Collier}]{liu2021fast}
Fangyu Liu, Ivan Vuli{\'c}, Anna Korhonen, and Nigel Collier. 2021.
\newblock Fast, effective, and self-supervised: Transforming masked language
  models into universal lexical and sentence encoders.
\newblock In \emph{EMNLP}.

\bibitem[{Liu et~al.(2019)Liu, Ott, Goyal, Du, Joshi, Chen, Levy, Lewis,
  Zettlemoyer, and Stoyanov}]{liu2019roberta}
Yinhan Liu, Myle Ott, Naman Goyal, Jingfei Du, Mandar Joshi, Danqi Chen, Omer
  Levy, Mike Lewis, Luke Zettlemoyer, and Veselin Stoyanov. 2019.
\newblock Ro{BERT}a: A robustly optimized {BERT} pretraining approach.
\newblock \emph{arXiv}.

\bibitem[{Logeswaran and Lee(2018)}]{logeswaran2018efficient}
Lajanugen Logeswaran and Honglak Lee. 2018.
\newblock An efficient framework for learning sentence representations.
\newblock In \emph{ICLR}.

\bibitem[{Marelli et~al.(2014)Marelli, Menini, Baroni, Bentivogli, Bernardi,
  and Zamparelli}]{marelli-etal-2014-sick}
Marco Marelli, Stefano Menini, Marco Baroni, Luisa Bentivogli, Raffaella
  Bernardi, and Roberto Zamparelli. 2014.
\newblock A {SICK} cure for the evaluation of compositional distributional
  semantic models.
\newblock In \emph{{LREC}}.

\bibitem[{Pang and Lee(2004)}]{pang2004sentimental}
Bo~Pang and Lillian Lee. 2004.
\newblock A sentimental education: Sentiment analysis using subjectivity
  summarization based on minimum cuts.
\newblock In \emph{ACL}.

\bibitem[{Pang and Lee(2005)}]{pang2005seeing}
Bo~Pang and Lillian Lee. 2005.
\newblock Seeing stars: Exploiting class relationships for sentiment
  categorization with respect to rating scales.
\newblock In \emph{ACL}.

\bibitem[{Raffel et~al.(2019)Raffel, Shazeer, Roberts, Lee, Narang, Matena,
  Zhou, Li, and Liu}]{2019t5}
Colin Raffel, Noam Shazeer, Adam Roberts, Katherine Lee, Sharan Narang, Michael
  Matena, Yanqi Zhou, Wei Li, and Peter~J. Liu. 2019.
\newblock Exploring the limits of transfer learning with a unified text-to-text
  transformer.
\newblock \emph{arXiv}.

\bibitem[{Reimers and Gurevych(2019)}]{reimers2019sentence}
Nils Reimers and Iryna Gurevych. 2019.
\newblock Sentence-bert: Sentence embeddings using siamese bert-networks.
\newblock In \emph{EMNLP-IJCNLP}.

\bibitem[{Socher et~al.(2013)Socher, Perelygin, Wu, Chuang, Manning, Ng, and
  Potts}]{socher2013recursive}
Richard Socher, Alex Perelygin, Jean Wu, Jason Chuang, Christopher~D Manning,
  Andrew~Y Ng, and Christopher Potts. 2013.
\newblock Recursive deep models for semantic compositionality over a sentiment
  treebank.
\newblock In \emph{EMNLP}.

\bibitem[{Voorhees and Tice(2000)}]{voorhees2000building}
Ellen~M Voorhees and Dawn~M Tice. 2000.
\newblock Building a question answering test collection.
\newblock In \emph{SIGIR}.

\bibitem[{Wang et~al.(2022)Wang, Li, Huang, Dou, Kong, and
  Shao}]{wang2022sncse}
Hao Wang, Yangguang Li, Zhen Huang, Yong Dou, Lingpeng Kong, and Jing Shao.
  2022.
\newblock Sncse: Contrastive learning for unsupervised sentence embedding with
  soft negative samples.
\newblock \emph{arXiv}.

\bibitem[{Wang and Isola(2020)}]{wang2020understanding}
Tongzhou Wang and Phillip Isola. 2020.
\newblock Understanding contrastive representation learning through alignment
  and uniformity on the hypersphere.
\newblock In \emph{ICML}.

\bibitem[{Wiebe et~al.(2005)Wiebe, Wilson, and Cardie}]{wiebe2005annotating}
Janyce Wiebe, Theresa Wilson, and Claire Cardie. 2005.
\newblock Annotating expressions of opinions and emotions in language.
\newblock \emph{Language resources and evaluation}.

\bibitem[{Wu et~al.(2021)Wu, Gao, Zang, Han, Wang, and Hu}]{wu2021esimcse}
Xing Wu, Chaochen Gao, Liangjun Zang, Jizhong Han, Zhongyuan Wang, and Songlin
  Hu. 2021.
\newblock Esimcse: Enhanced sample building method for contrastive learning of
  unsupervised sentence embedding.
\newblock \emph{arXiv}.

\bibitem[{Yan et~al.(2021)Yan, Li, Wang, Zhang, Wu, and Xu}]{yan2021consert}
Yuanmeng Yan, Rumei Li, Sirui Wang, Fuzheng Zhang, Wei Wu, and Weiran Xu. 2021.
\newblock Consert: A contrastive framework for self-supervised sentence
  representation transfer.
\newblock In \emph{ACL}.

\bibitem[{Zhang et~al.(2021)Zhang, He, Liu, Bing, and
  Li}]{zhang2021bootstrapped}
Yan Zhang, Ruidan He, Zuozhu Liu, Lidong Bing, and Haizhou Li. 2021.
\newblock Bootstrapped unsupervised sentence representation learning.
\newblock In \emph{ACL}.

\bibitem[{Zhou et~al.(2022)Zhou, Zhang, Zhao, and
  Wen}]{zhou-etal-2022-debiased}
Kun Zhou, Beichen Zhang, Xin Zhao, and Ji-Rong Wen. 2022.
\newblock Debiased contrastive learning of unsupervised sentence
  representations.
\newblock In \emph{ACL}.

\bibitem[{Zhu et~al.(2015)Zhu, Kiros, Zemel, Salakhutdinov, Urtasun, Torralba,
  and Fidler}]{Zhu_2015_ICCV}
Yukun Zhu, Ryan Kiros, Rich Zemel, Ruslan Salakhutdinov, Raquel Urtasun,
  Antonio Torralba, and Sanja Fidler. 2015.
\newblock Aligning books and movies: Towards story-like visual explanations by
  watching movies and reading books.
\newblock In \emph{ICCV}.

\end{thebibliography}
\bibliographystyle{acl_natbib}

\appendix

\section{Appendix}\label{sec:appendix}
\subsection{The connection between the normalization function $g$ and Spearman’s Rank Correlation} \label{sec:appendix_rho}
The Spearman’s rank correlation of two lists of variables, $u=<u_1, ..., u_n>$ and $v=<v_1, ..., v_n>$, is the Pearson correlation coefficient, $\rho$, of their ranks, $r^u$ and $r^v$, as follows:

\begin{equation}
        \rho(r^u, r^v) = \frac{\sum_{i=1}^n\big\{\frac{1}{n}(r^u_i-\overline{r}^u)\times(r^v_i-\overline{r}^v)\big\}}{\sigma(r^u) \times \sigma(r^v)},
\end{equation}

where $\overline{r}^u$ and $\overline{r}^v$ are the mean of the rank variables, $\sigma(r)^u$ and $\sigma(r)^v$ are the standard deviations of ranks. 
Then, this can be re-written as follows:

\begin{equation}
    \begin{split}
    &\rho(r^u, r^v) =\\
    &\big(\frac{1}{\sqrt{n}}(r^u-\overline{r}^u) / \sigma(r^u)\big)^{\intercal}
    \big(\frac{1}{\sqrt{n}}(r^v-\overline{r}^v)/\sigma(r^v)\big).
    \end{split}
\end{equation}

Thus, the inner product of the two rank vectors after normalization with g is equivalent to the Spearman’s rank correlation of the rank variables.

\begin{table*}[t]
\centering
\begin{tabular}{@{}lcccccccc@{}}
\toprule
Model                     & MR             & CR             & SUBJ           & MPQA           & SST            & TREC           & MRPC           & AVG            \\ \midrule
SimCSE                    & 81.18          & 86.46          & 94.45          & 88.88          & 85.50          & 89.80          & 74.43          & 85.81          \\
SimCSE w/ MLM             & \textbf{82.92} & 87.23          & \textbf{95.71} & 88.73          & \textbf{86.81} & 87.01          & \textbf{78.07} & 86.64          \\
RankEncoder-SimCSE w/ MLM & 82.14          & \textbf{87.31} & 95.35          & \textbf{89.05} & 86.66          & \textbf{91.00} & 76.06          & \textbf{86.80} \\ \bottomrule
\end{tabular}
\caption{Transfer task results of baselines and RankEncoder. 
We use RankEncoder with base encoder SimCSE.
MLM represents that the model is trained by both loss functions: its loss function and the masked language modeling loss used in pre-trained language models such as BERT.
We set the weight parameter of the MLM loss function to 0.1.}
\label{tab:transfer_task}
\end{table*}

\subsection{$\lambda_{\text{train}}$ Analysis}\label{sec:lmb_analysis}
\begin{figure}[t]
    \centering
    \includegraphics[width=0.9\linewidth]{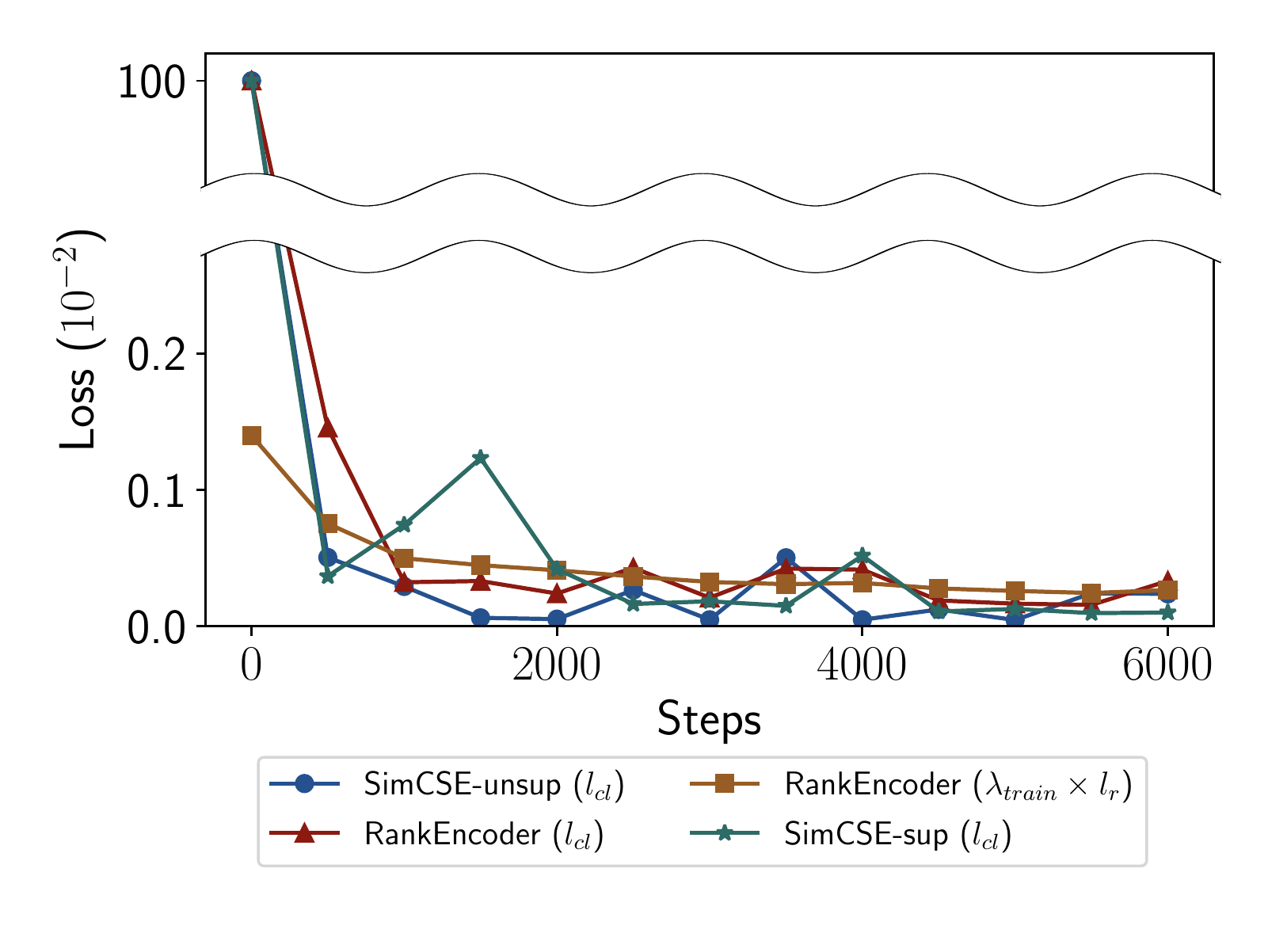}
    \caption{The training loss curves of SimCSE and RankEncoder.
    X-axis represents a training step, and Y-axis is a scaled loss.
    After few training steps, the three losses converge in a similar value.
    Setting $\lambda_{\text{train}}$ to a small value, $0.05$, results in similar weights on the two loss functions of RankEncoder while maintaining the loss curve of the base encoder.}
    \label{fig:loss_plot}
\end{figure}

The RankEncoder loss, $l_r$, brings a large effect to RankEncoder's re-training process even when the weight parameter, $\lambda_{\text{train}}$, is set to a small value.
In this section, we show that the two losses, $l_{\text{cl}}$ and $l_r$, similarly affect to the total loss, $l_{\text{total}}$ in Eq. \ref{eq:loss}, when $\lambda_{\text{train}}=0.05$, which is the default setting we use for all experiments in this paper.
Figure \ref{fig:loss_plot} shows the training loss curves of RankEncoder and SimCSE-unsup with the same random seed.
We show the two losses, $l_{\text{cl}}$ and $l_r$, of RankEncoder separately.
SimCSE-unsup’s loss rapidly decreases at the beginning, and converges to a value less than 0.001.
We see a similar pattern in the contrastive loss of RankEncoder, which is the same loss function as SimCSE-unsup.
In contrast, $\lambda_{\text{train}} \times l_r$ starts from a much lower value than $l_{\text{cl}}$; even without the weight parameter, $l_r$ is still much lower than $l_{\text{cl}}$.
After few training steps, $\lambda_{\text{train}} \times l_r$ converges close to the value of $l_{\text{cl}}$.
Given that $\lambda_{\text{train}}$ determines the scale of two losses of our hinge loss function (Eq. \ref{eq:loss}), we expect that increasing $\lambda_{\text{train}}$ brings RankEncoder's loss curve converged to higher than SimCSE's loss.
This result shows that $\lambda_{\text{train}}=0.05$ is optimal value that maintaining the RankEncoder's loss curve similar to the base encoder's loss curve, while balancing the weights of the two losses, $l_{\text{cl}}$ and $l_{r}$.

\begin{table*}[t]
\centering
\begin{tabular}{@{}lcccccccc@{}}
\toprule
{Model}         & {STS12}                             & {STS13}                             & {STS14}                             & {STS15}                             & {STS16}                             & {STS-B}                             & {SICK-R}                            & {AVG}                               \\ \midrule
SimCSE        & { 68.1}$\scriptstyle\pm${\scriptsize 3.3} & { 81.4}$\scriptstyle\pm${\scriptsize 1.6} & { 73.8}$\scriptstyle\pm${\scriptsize 2.4} & { 81.8}$\scriptstyle\pm${\scriptsize 1.4} & { 78.3}$\scriptstyle\pm${\scriptsize 0.6} & { 77.3}$\scriptstyle\pm${\scriptsize 2.3} & { 71.0}$\scriptstyle\pm${\scriptsize 0.4} & { 76.0}$\scriptstyle\pm${\scriptsize 1.5} \\
\rowcolor[HTML]{EFEFEF} 
+ {\small RankEncoder} & \bf{{ 75.0}}$\scriptstyle\pm${\scriptsize 0.6} & \bf{{ 82.0}}$\scriptstyle\pm${\scriptsize0.7}  & \bf{{ 75.2}}$\scriptstyle\pm${\scriptsize 0.2} & \bf{{ 83.0}}$\scriptstyle\pm${\scriptsize 0.1} & \bf{{ 79.8}}$\scriptstyle\pm${\scriptsize 0.1} & \bf{{ 80.4}}$\scriptstyle\pm${\scriptsize 0.6} & \bf{{ 71.1}}$\scriptstyle\pm${\scriptsize 1.2} & \bf{{ 78.1}}$\scriptstyle\pm${\scriptsize 0.1} \\ \midrule
PromptBERT    & { 72.1}$\scriptstyle\pm${\scriptsize 0.2} & { 84.6}$\scriptstyle\pm${\scriptsize 0.3} & { 76.8}$\scriptstyle\pm${\scriptsize 0.1} & { 84.2}$\scriptstyle\pm${\scriptsize 0.3} & { 80.4}$\scriptstyle\pm${\scriptsize 0.3} & { 81.8}$\scriptstyle\pm${\scriptsize 0.3} & { 69.5}$\scriptstyle\pm${\scriptsize 0.2} & { 78.5}$\scriptstyle\pm${\scriptsize 0.0} \\
\rowcolor[HTML]{EFEFEF} 
+ {\small RankEncoder} & \bf{{ 74.2}}$\scriptstyle\pm${\scriptsize 0.3} & \bf{{ 85.2}}$\scriptstyle\pm${\scriptsize 0.2} & \bf{{ 77.7}}$\scriptstyle\pm${\scriptsize 0.2} & \bf{{ 84.4}}$\scriptstyle\pm${\scriptsize 0.3} & \bf{{ 80.7}}$\scriptstyle\pm${\scriptsize 0.5} & \bf{{ 82.1}}$\scriptstyle\pm${\scriptsize 0.4} & \bf{{ 71.2}}$\scriptstyle\pm${\scriptsize 0.2} & \bf{{ 79.4}}$\scriptstyle\pm${\scriptsize 0.2} \\ \midrule
SNCSE         & { 70.2}$\scriptstyle\pm${\scriptsize 0.5} & { 84.1}$\scriptstyle\pm${\scriptsize 0.5} & { 77.1}$\scriptstyle\pm${\scriptsize 0.4} & \bf{{ 83.2}}$\scriptstyle\pm${\scriptsize 0.5} & { 80.7}$\scriptstyle\pm${\scriptsize 0.1} & { 80.7}$\scriptstyle\pm${\scriptsize 0.6} & { 75.0}$\scriptstyle\pm${\scriptsize 0.1} & { 78.7}$\scriptstyle\pm${\scriptsize 0.3} \\
\rowcolor[HTML]{EFEFEF} 
+ {\small RankEncoder} & \bf{{ 73.9}}$\scriptstyle\pm${\scriptsize 0.6} & \bf{{ 84.5}}$\scriptstyle\pm${\scriptsize 0.5} & \bf{{ 78.0}}$\scriptstyle\pm${\scriptsize 0.3} & { 83.0}$\scriptstyle\pm${\scriptsize 0.5} & \bf{{ 81.0}}$\scriptstyle\pm${\scriptsize 0.2} & \bf{{ 81.2}}$\scriptstyle\pm${\scriptsize 0.2} & \bf{{ 75.3}}$\scriptstyle\pm${\scriptsize 0.1} & \bf{{ 79.6}}$\scriptstyle\pm${\scriptsize 0.2} \\ \bottomrule
\end{tabular}
\caption{Semantic textual similarity performance of sentence encoders.
We measure Spearman's rank correlation between the human-annotated scores and the model's predictions. We report the mean performance and standard deviation of three separate trials with different random seeds.}
\label{tab:inc}
\end{table*}

\begin{table*}[t]
\centering
\begin{tabular}{@{}lccc@{}}
\toprule
                                                & \multicolumn{3}{c}{Base Encoder $E_1$} \\
                                                & SimCSE     & PromptBERT     & SNCSE    \\ \midrule
$E_1$                                           & 44.59      & 49.56          & 48.19    \\
$\text{RankEncoder}_{E_1}$                      & 46.73      & 50.06          & 49.44    \\
$\text{RankEncoder}_{E_1}-\text{retrain}$ & 48.41      & 50.75          & 49.80    \\
$\text{RankEncoder}_{E_1}-\text{retrain}-\text{inf}$               & 48.73      & 50.93          & 49.92    \\ \bottomrule
\end{tabular}
\caption{Semantic textual similarity performance of variations of RankEncoder. 
$E$ is the base encoder. $\text{RankEncoder}_{E}$ is RankEncoder with Eq. \ref{eq:rank_vector}. $\text{RankEncoder}-\text{retrain}$ is RankEncoder with Eq. \ref{eq:loss}. $\text{RankEncoder}-\text{retrain}-\text{inf}$ is RankEncoder with Eq. \ref{eq:inference}.}
\label{tab:inc_pos}
\end{table*}

\begin{figure*}[t]
    \begin{subfigure}{.32\textwidth}
        \centering
        \includegraphics[width=1.\linewidth]{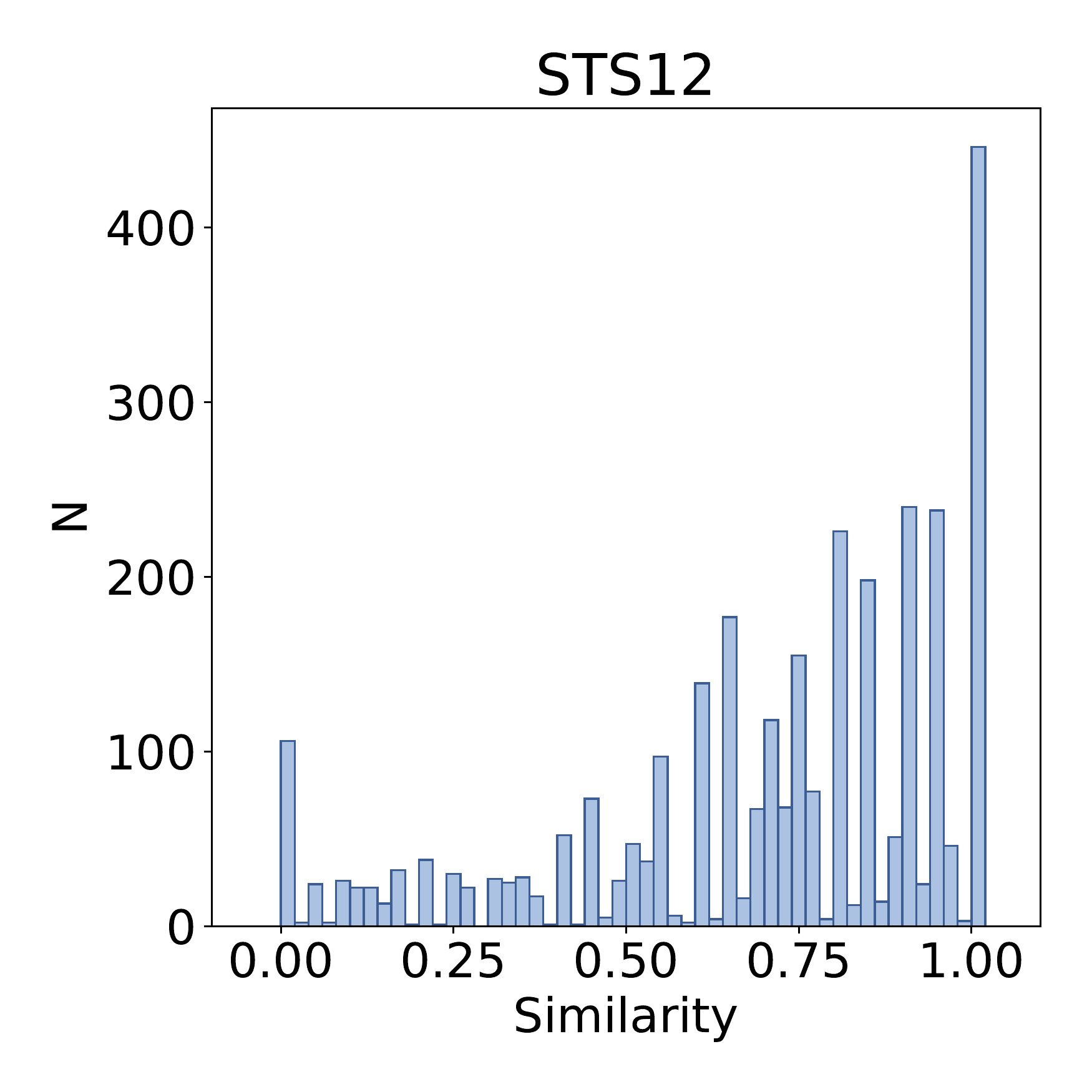}
    \end{subfigure}
    \begin{subfigure}{.32\textwidth}
        \centering
        \includegraphics[width=1.\linewidth]{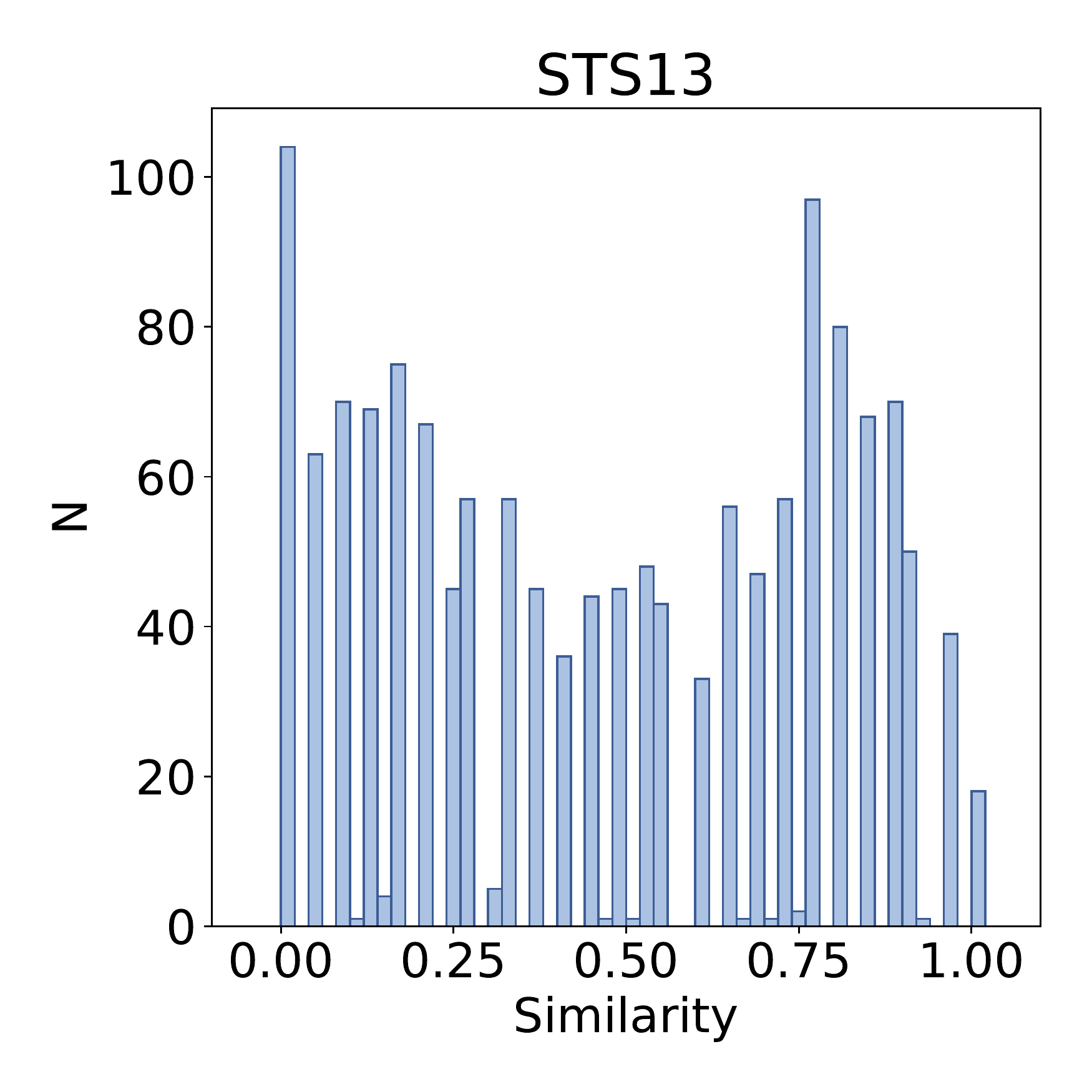}
    \end{subfigure}
    \begin{subfigure}{.32\textwidth}
        \centering
        \includegraphics[width=1.\linewidth]{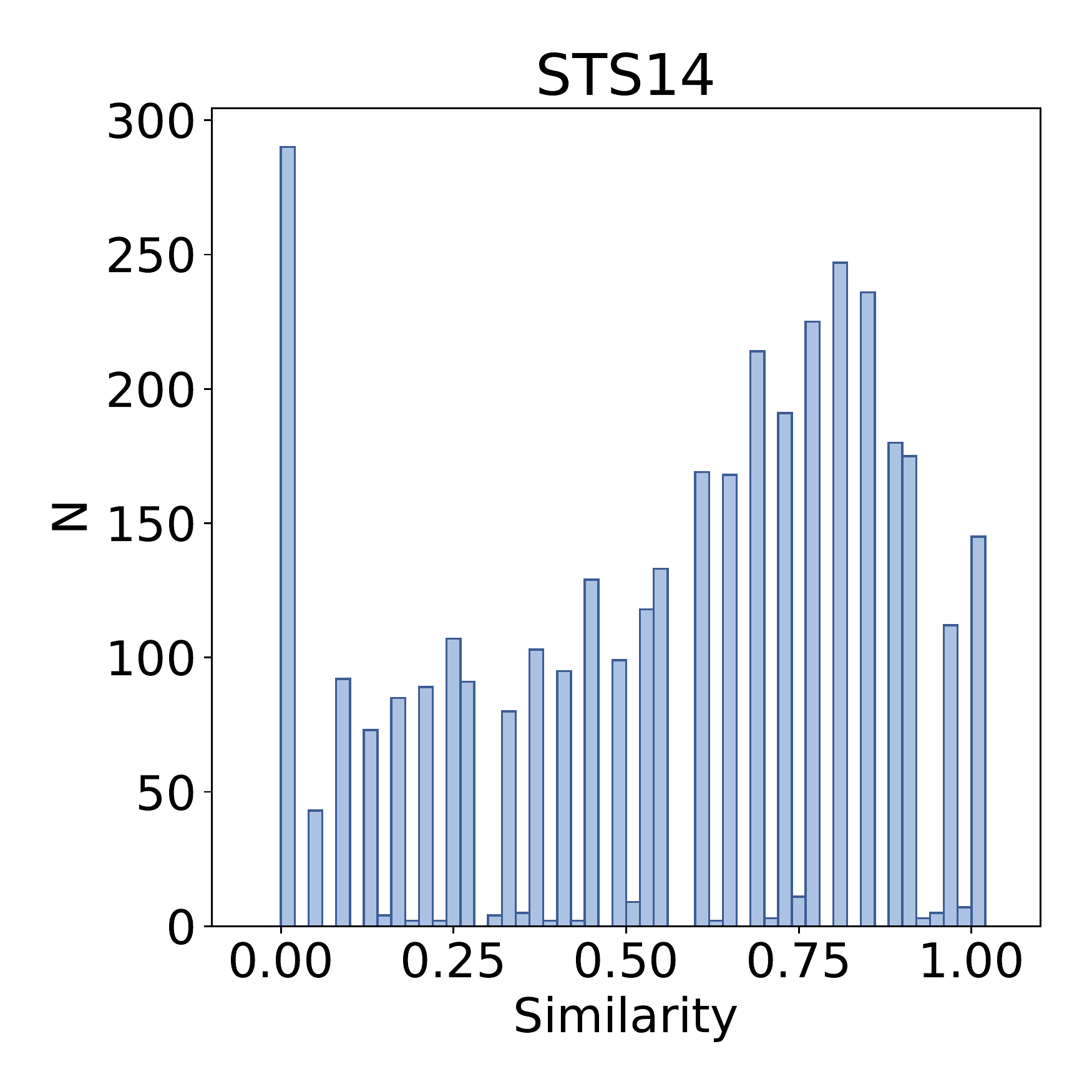}
    \end{subfigure}
    \begin{subfigure}{.32\textwidth}
        \centering
        \includegraphics[width=1.\linewidth]{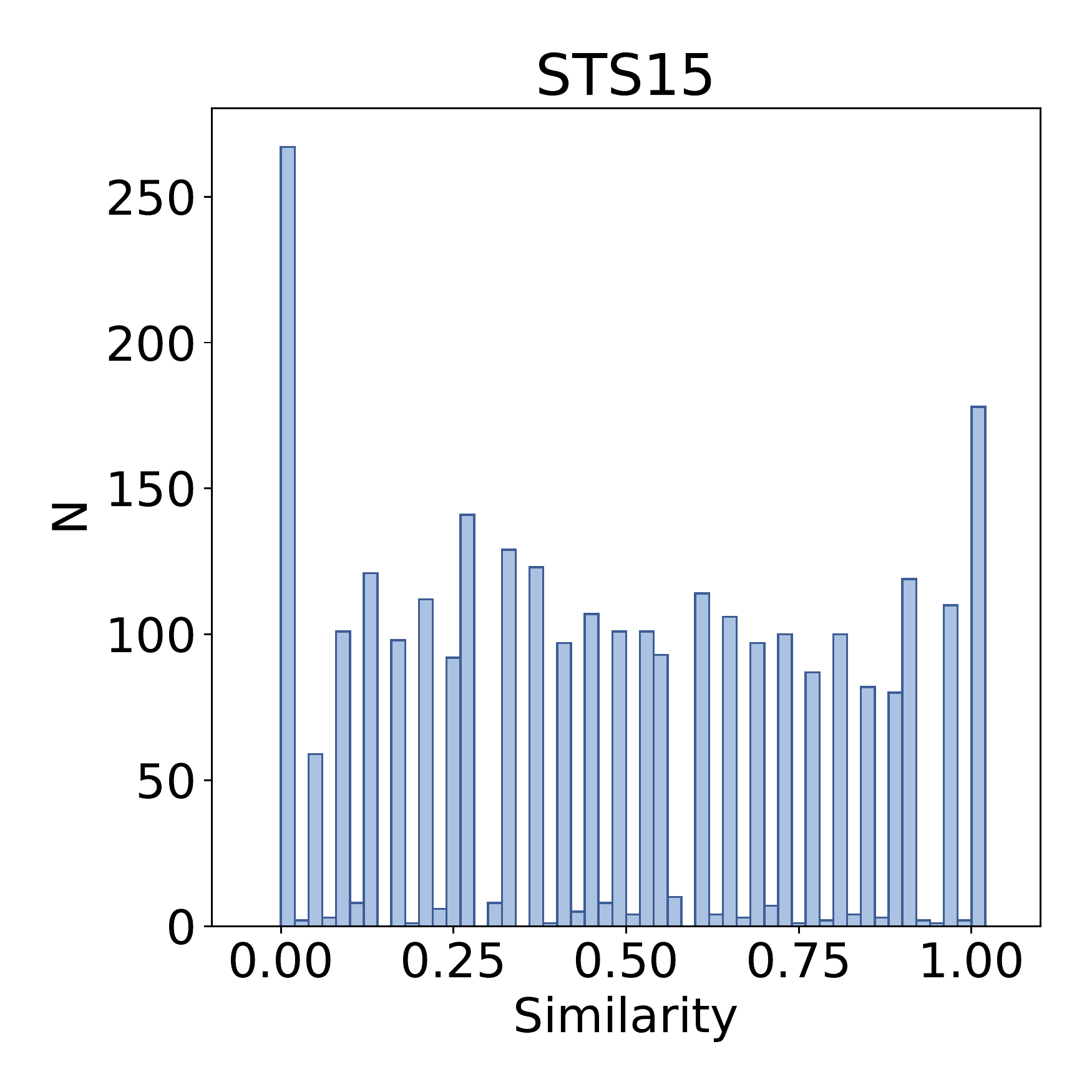}
    \end{subfigure}
    \begin{subfigure}{.32\textwidth}
        \centering
        \includegraphics[width=1.\linewidth]{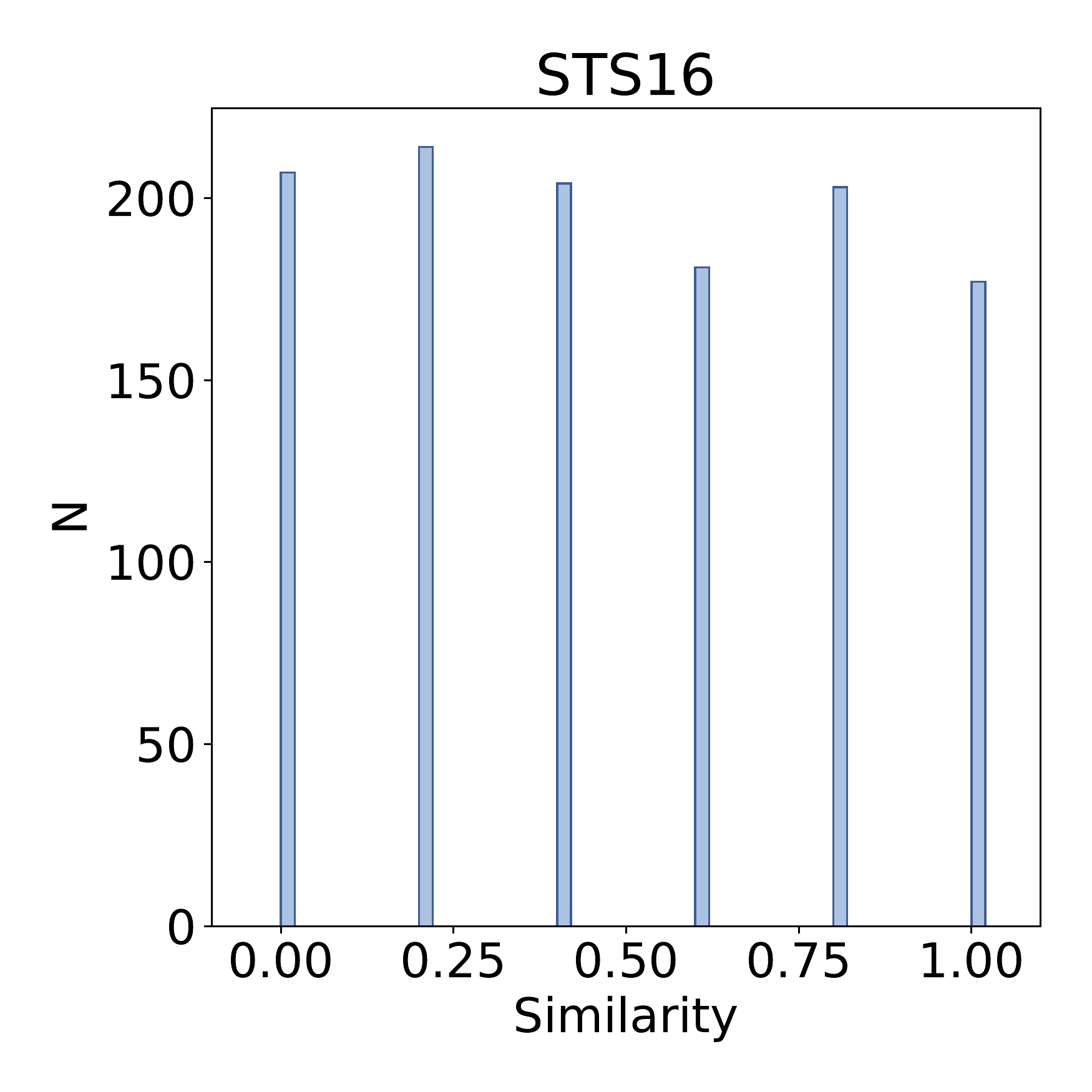}
    \end{subfigure}
    \begin{subfigure}{.32\textwidth}
        \centering
        \includegraphics[width=1.\linewidth]{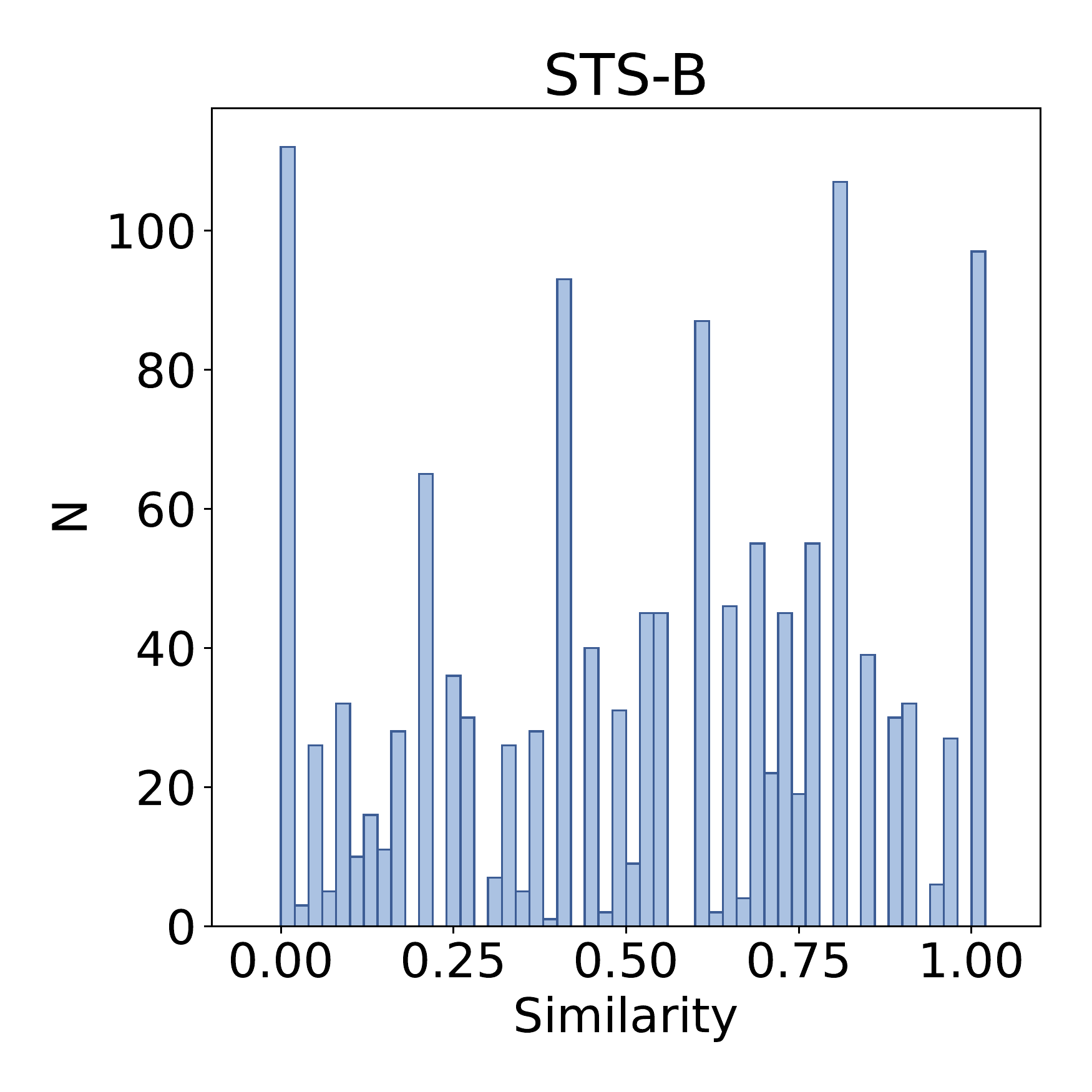}
    \end{subfigure}
    \begin{subfigure}{.32\textwidth}            
        \includegraphics[width=1.\linewidth]{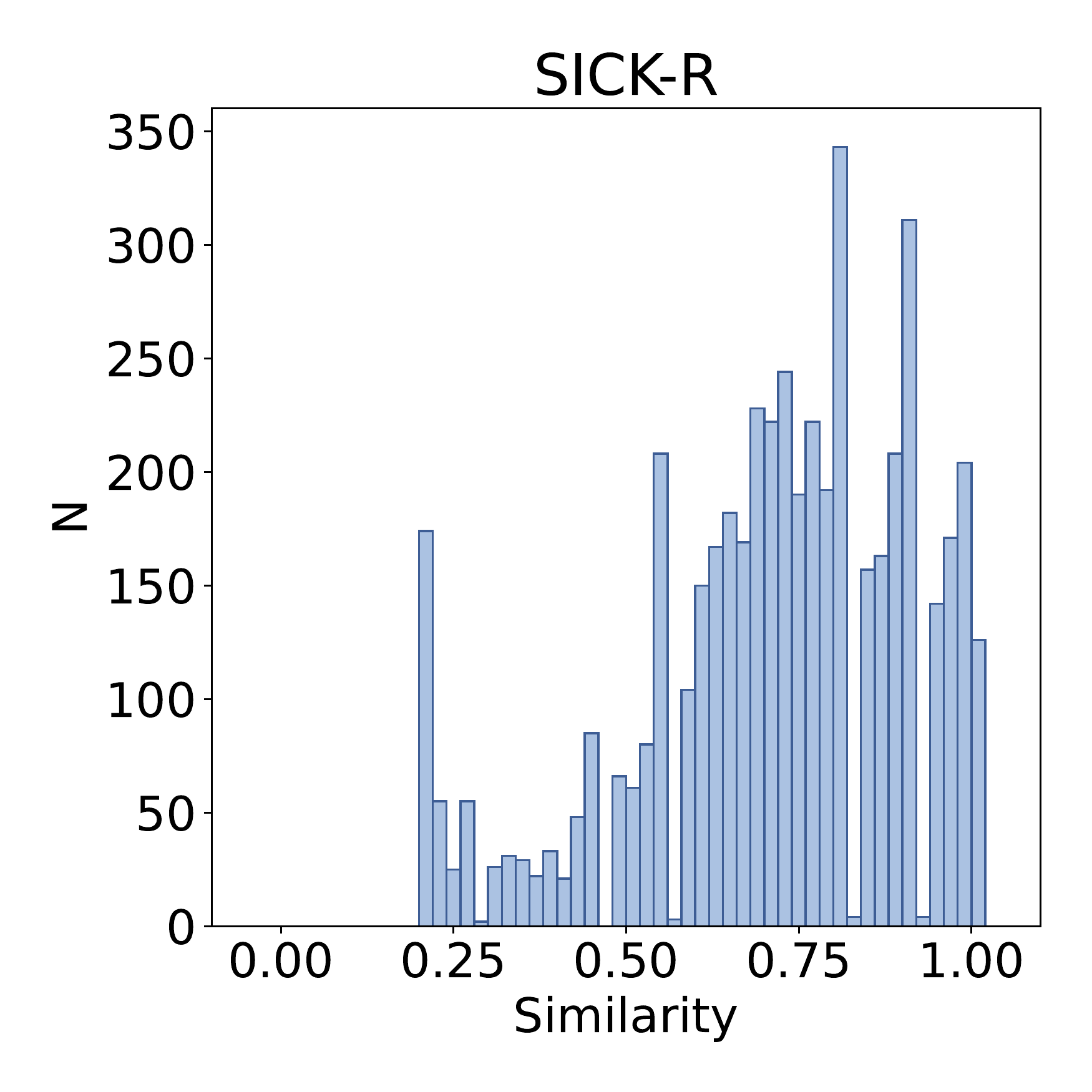}
    \end{subfigure}
    \caption{Similarity distributions of semantic textual similarity benchmark datasets. We scale the similarity scores between 0.0 and 1.0.}
    \label{fig:sim_histogram}
    \vspace{-1em}
\end{figure*}

The loss curve of a supervised sentence encoder provides a reference point for comparison between the loss curves of unsupervised sentence encoders.
In Figure \ref{fig:loss_plot}, all unsupervised sentence encoders' loss curves show a rapidly decreasing pattern, which implies overfitting in training.
To verify whether this pattern comes from unsupervised training, we show the loss curve of the supervised sentence encoder, SimCSE-sup, in Figure \ref{fig:loss_plot}.
In this experiment, we measure the same contrastive loss used in unsupervised sentence encoders but in the SimCSE-sup's fully supervised training process.
We see the same pattern also holds for SimCSE-sup and verify that the rapidly decreasing pattern is not the problem that only occurs in unsupervised training.

\subsection{Similarity Distribution of STS Benchmark Datasets}\label{sec:appendix_sim_dist}
Semantic textual similarity datasets have different similarity distributions.
Since RankEncoder is specifically effective for similar sentence pairs, we expect that RankEncoder brings a more performance increase on datasets with more similar sentence pairs.
We show the similarity distribution of each STS dataset in Figure \ref{fig:sim_histogram}.
In this figure, we normalize the similarity scores between $0$ and $1$.
The result shows that the similarity distributions of STS12, STS14, and SICK-R are skewed to a high similarity score and STS13's similarity distribution has a distinct peak at a high similarity score.
From the results in Table \ref{tab:sts}, we see that RankEncoder is more effective on STS12, STS13, STS14, SICK-R, and show the relation between the performance increase and the similarity distribution of each dataset.

\subsection{Transfer Tasks}
\yeon{We verify that applying our approach to an existing unsupervised sentence encoder increases the performance on transfer tasks.
We use the following seven transfer tasks to evaluate sentence embeddings:
MR~\citep{pang2005seeing}, CR~\citep{hu2004mining}, SUBJ~\citep{pang2004sentimental}, MPQA~\citep{wiebe2005annotating}, SST~\citep{socher2013recursive}, TREC~\citep{voorhees2000building}, and MRPC~\citep{dolan2005automatically}.
These transfer tasks employ an additional single-layer neural network to transform sentence embeddings into the appropriate output format for a given task.
The single-layer neural network is trained with the training set of each task.
We use the SentEval toolkit~\cite{conneau2018senteval} for evaluation.
Table \ref{tab:transfer_task} shows the performance of SimCSE and RankEncoder on these transfer tasks.
We use SimCSE as the base encoder of RankEncoder.
Recently, SimCSE~\citep{gao-etal-2021-simcse} has shown that training sentence encoders with auxiliary masked language modeling (MLM) loss enhances their performance on transfer tasks.
Inspired by this finding, we use MLM loss when training RankEncoder.
The experimental results show that our approach increases the average performance on transfer tasks by 0.16\%p.
This performance gain is relatively small when we compare it with the performance gain on STS benchmark datasets shown in Table \ref{tab:sts}.
This is because the sentence embedding quality is not directly connected to the objective of transfer tasks~\citep{gao-etal-2021-simcse}.
}

\subsection{Universality of RankEncoder}\label{sec:appendix_universality}
In this section, we report the detailed experimental results of Figure \ref{fig:inc}.
Table \ref{tab:inc} shows the results.

\subsection{The performance of RankEncoder on Similar Sentence Pairs}\label{sec:appendix_pef_similar_pairs}
We report the detailed results of Figure \ref{fig:perf_similar_pairs} in Table \ref{tab:inc_pos}.

\subsection{Computational Cost} \label{appendix:computation_cost}
\revisionyeon{In this section, we describe the details of the computational efficiency of RankEncoder.}
\paragraph{Pre-Computation:}
\revisionyeon{We pre-compute sentence vectors of corpus $\mathcal{C}$ for training and inference. This takes a few seconds on a single V100 GPU.}
\paragraph{Training:}
\revisionyeon{Most of the additional training time comes from calculating a rank vector similarity matrix (the matrix in Figure \ref{fig:method}).}
\revisionyeon{First, we calculate a rank vector for every sentence in a given batch. The time complexity of calculating a rank vector is $O(N\times D)$, where $N$ is the number of pre-indexed vectors, and $D$ is the dimension of the vectors. Assuming a batch size of $B$, the time complexity of this step is $O(B\times N \times D)$.}
\revisionyeon{Second, we calculate a $B \times B$ rank vector similarity matrix. This is $O(B \times B \times N)$ since the dimension size of the rank vector is $N$.}
\revisionyeon{The total time complexity is $O(B \times D \times N + B \times B \times N)$, which is $O(B \times D \times N)$, assuming the batch size is much smaller than the dimension size.}
\revisionyeon{As a result, the total training time is 1.5 hours, 0.5 hours (the base encoder’s training time) + 1.0 hours (the additional training time brought by our approach).}
\paragraph{Inference:}
\revisionyeon{RankEncoder’s inference process comprises two steps: 1) predicting the sentence vector of an input sentence and 2) computing similarity scores between the input sentence vector and the pre-indexed vectors; we exclude the indexing time since indexing is completed before inference.}
\revisionyeon{The first step takes the same inference time as BERT-base (0.07 seconds for a given sentence on a single V100 GPU) as RankEncoder uses BERT-base.}
\revisionyeon{The second step entails matrix multiplication of an $N \times D$ matrix and a $D \times 1$ matrix (this takes 0.0013 seconds), which takes 1.8\% of the whole inference time.}
\revisionyeon{Thus, our method increases the inference time by 1.8\%.}

\end{document}